\documentclass[final,5p,times,twocolumn]{elsarticle}

\usepackage{amssymb}
\usepackage{lipsum}
\usepackage{orcidlink}
\usepackage{booktabs}
\usepackage{amsmath}
\usepackage{cleveref}
\usepackage{forest}
\usepackage{caption}
\usepackage{float}
\usepackage{enumitem}
\usepackage{subcaption}

\journal{High Energy Astrophysics}

\begin{document}

\begin{frontmatter}



\title{Impact of Evidence Theory Uncertainty on Training Object Detection Models}


\author[first]{M. Tahasanul Ibrahim\orcidlink{0000-0002-7468-0171}}
\author[first]{Rifshu Hussain Shaik}
\author[first]{Andreas Schwung\orcidlink{0000-0001-8405-0977}}
\affiliation[first]{organization={Department of Automation Technology and Learning Systems, South Westphalia University of Applied Sciences},
            addressline={Lübecker Ring 2}, 
            city={Soest},
            postcode={59494}, 
            state={North Rhine-Westphalia},
            country={Germany}}

\begin{abstract}
This paper investigates the use of Evidence Theory to enhance the training efficiency of object detection models by incorporating uncertainty into the feedback loop. In each training iteration, during the validation phase, Evidence Theory is applied to establish a relationship between ground truth labels and predictions. The Dempster-Shafer rule of combination is used to quantify uncertainty based on the evidence from these predictions. This uncertainty measure is then utilized to weight the feedback loss for the subsequent iteration, allowing the model to adjust its learning dynamically. By experimenting with various uncertainty-weighting strategies, this study aims to determine the most effective method for optimizing feedback to accelerate the training process. The results demonstrate that using uncertainty-based feedback not only reduces training time but can also enhance model performance compared to traditional approaches. This research offers insights into the role of uncertainty in improving machine learning workflows, particularly in object detection, and suggests broader applications for uncertainty-driven training across other AI disciplines.
\end{abstract}




\end{frontmatter}




\section{Introduction}
Object detection is a fundamental challenge in the field of computer vision, focusing on identifying and classifying object instances from a wide range of predefined categories within images. This area has experienced transformative growth, driven by advancements in deep convolutional neural networks (CNNs). These advancements have led to the development of sophisticated object detection frameworks, including both one-stage and two-stage detectors, which use classifiers to distinguish objects from the background and regressors to predict precise bounding boxes. A critical component of this process is the use of ground truth boxes, which define classification tasks by assigning positive and negative labels during training, often based on Intersection Over Union (IOU) thresholds \cite{8917599}.

In this evolving landscape, our research introduces a method that applies the Dempster-Shafer (D-S) theory of evidence to enhance the training of object detection models by incorporating uncertainty. This approach centers on Dempster's Rule of Combination, a principle within D-S theory designed to intelligently integrate predictions from different object detection models. Unlike traditional fusion methods that rely on averaging or selecting the most confident prediction, Dempster's Rule evaluates the degree of consensus and disagreement among model predictions, offering a more refined and effective approach to decision-making.

Our method further leverages the uncertainty captured through D-S fusion in the training phase, using weighted losses mapping in the feedbackward loop. This integration of fusion uncertainty alongside the loss values enables a more adaptive optimization process, enhancing model robustness, and faster reaching to the local minima. 

This approach is particularly beneficial in high-stakes applications such as autonomous vehicle navigation, security surveillance, and medical diagnostics, where accurate object detection is critical. By treating each model's output as a piece of evidence, complete with its own belief and plausibility measures, the D-S theory-based fusion significantly improves detection performance, especially in cases of overlapping or ambiguous predictions where conventional methods may fail.

Moreover, our research explores the ability of this approach to reduce false positives, thereby increasing the overall reliability of object detection systems. By applying D-S theory in decision-making, we reduce the likelihood of incorrect detections, enhancing the system's trustworthiness.

This novel application of Dempster-Shafer theory for merging model predictions marks a significant advancement in the field of object detection. By integrating classical decision-making theories with modern computational models, it opens new avenues for improving detection accuracy and reliability \cite{Shafer1976, NeagoeGhenea2022, SunSong2021, HuoMartinez2022, DeVilliersLaskey2015}. This contribution addresses critical challenges in uncertainty and multi-model predictions while setting a foundation for future research in the field.

\begin{table}[ht]    
	\centering
	\caption{List of symbols, abbreviations, and acronyms.}
    \resizebox{\columnwidth}{!}{
	\begin{tabular}{l l}
		\toprule
		\textbf{Symbol} & \textbf{Description} \\ 		
		\midrule
            DSET            & Dempster-Shafer evidence theory\\
            DSRC            & Dempster-Shafer rule of combination\\
            EC              & Ensemble classifier\\
            ET              & Evidence theory\\
            UQ              & Uncertainty quantification\\
            YOLO            & You Only Look Once\\
            DL              & Dynamic Loss\\
            INFUSION        & Adaptive information fusion using ET and UQ\\
            DIU             & Direct Injection of Uncertainty\\
            AIU             & Average Injection of Uncertainty\\
            BPA             & Basic Probability Assignment\\
            IOU             & Intersection Over Union\\
            RNN             & Recurrent Neural Networks\\
            mAP             & Mean Average Precision\\
            CARL            & Classification Aware Regression Loss\\
            RLO             & Reinforcement Learning Optimization \\
            OHEM            & Online Hard Example Mining\\
            GHM             & Gradient Harmonizing Mechanism\\
            AVW             & Adaptive Variance Weighting\\
            CNN             & Convolutional Neural Networks\\
            RPN             & Region Proposal Network\\
            FC Layers       & Fully Connected layers\\
            RoI             & Regions of Interest\\
            Faster R-CNN    & Faster Region-based Convolutional Neural Network\\
            $Pl$            & Plausibility\\
            $Bel$           & Belief Function\\
            $K$             & Uncertainty\\
            $w(K)$          & Multiplication Factor\\
            $\Phi$          & Certainty\\
            $\Theta$        & Subsets\\
            $\hat\Theta$    & Revised Subsets\\
            $E_{x}$         & Evidence\\
            $L_{x}$         & Predicted Label\\
            $S_{x}$         & Predicted Label Score\\
            $\hat E_{x}$    & Normalized Evidence\\
            $\hat L_{x}$    & Normalized Predicted Label\\
            $\hat S_{x}$    & Normalized Predicted Label Score\\
		\bottomrule
	\end{tabular}
    }
	\label{table__list_symbols} 
\end{table}

\subsection{Key Contributions}
Here, we outline the primary innovations introduced by our work in enhancing object detection through uncertainty quantification. These contributions aim to bridge classical decision-making frameworks with modern machine learning techniques, offering a comprehensive approach to improving model accuracy and training efficiency.

\begin{itemize}
    \item \textbf{Novel Integration of Evidence Theory with Object Detection Models}: We propose a new approach that quantifies uncertainty during model training using Evidence Theory, enhancing the object detection process.
    
    \item \textbf{Uncertainty-Weighted Feedback Loss}: We introduce a method to dynamically weight feedback loss based on uncertainty computed via the Dempster-Shafer rule, improving model training efficiency.
    
    \item \textbf{Exploration of Multiple Weighting Strategies}: We investigate various strategies for applying uncertainty-weighted feedback loss, providing insights into the most effective methods for enhancing training efficiency in object detection algorithms like Faster R-CNN.
    
    \item \textbf{Acceleration of Model Training}: Our proposed uncertainty-based method significantly reduces training time while maintaining or improving model performance compared to traditional approaches.
\end{itemize}

This research sets the stage for future studies in object detection by combining classical decision-making theories with modern computational models. It provides a significant advancement in addressing challenges related to uncertainty and multi-model predictions while paving the way for innovations in improving the accuracy and reliability of object detection systems.

\section{Related Work}
    \subsection{Object Detection and Decision Assistance Systems}
    As object detection research has advanced, challenges related to imbalance problems have become increasingly prominent \cite{pang2019libra}. Imbalance typically refers to situations where the distribution of object classes in a dataset is uneven, making it difficult to train a model that performs consistently well across all classes. The most common imbalance issue is the foreground-to-background imbalance. Simultaneously training classification and regression tasks can also lead to objective imbalance, a domain that has received relatively little attention in object detection research \cite{luo2021dynamic}.
    
    To address these challenges, researchers have developed novel detector architectures and made improvements to the training pipeline. For instance, Libra R-CNN \cite{pang2019libra} focuses on creating a well-balanced feature pyramid, ensuring that multi-scale features receive equal attention. Later, AugFPN \cite{Guo_2020_CVPR} enhanced feature representation by fully extracting context features at multiple scales. CE-FPN \cite{luo2021cefpn} further improves performance by enhancing a series of feature channels, though this approach introduces computational complexity.
    
    Dynamic R-CNN \cite{zhang2020dynamic} leverages dynamic label assignment and Dynamic Smooth L1 loss to improve object detection proposals. This architecture effectively addresses data scarcity during early training while benefiting from high IOU training. By focusing on different aspects of the detector, these modules collectively enhance object detection efficiency.
    
    Beyond architecture improvements, addressing imbalance during the training phase is critical for object detection \cite{pang2019libra}. Online Hard Example Mining (OHEM) \cite{shrivastava2016training} automatically selects hard samples based on confidence, helping to balance foreground and background samples. IOU-based sampling \cite{pang2019libra} considers sample complexity based on IOU values, optimizing memory and efficiency. Focal Loss \cite{8237586} dynamically assigns higher weights to hard examples, and Dynamic Loss (DL) \cite{Zhao2018}, an improved version of focal loss, efficiently scales the traditional loss during training using a second-order term. DL, when implemented in YOLO-V2, achieved a mean average precision (mAP) of 88.51\%, a notable improvement over traditional loss functions \cite{Zhao2018}.
    
    The Gradient Harmonizing Mechanism (GHM) \cite{10.1609/aaai.v33i01.33018577} penalizes samples with similar gradients in one-stage detectors, addressing imbalance. A common solution to objective imbalance is to weight both the regression and classification tasks differently \cite{9042296}. To address regression loss, Classification Aware Regression Loss (CARL) \cite{Cao_2020_CVPR} extracts correlations between these tasks. KL-Loss \cite{he2019bounding}, inspired by multi-task learning methods \cite{8578879, Guo_2018_ECCV}, reduces the regression loss of uncertain samples through a trainable weight.
    
    The adaptive variance weighting (AVW) method, introduced by Luo et al. \cite{luo2021dynamic}, addresses imbalance in multi-scale loss by calculating the statistical variance of training loss at each step to determine importance. Based on this, the contributed loss is weighted during training. A novel approach called Reinforcement Learning Optimization (RLO) was also proposed to further optimize multi-scale loss, dynamically selecting the best scheme for different phases of training.
    
    This body of work is relevant to our research as it explores dynamic methods for optimizing multi-scale training loss by leveraging uncertainty through a weighted loss approach in the training feedback loop. By incorporating the uncertainty derived from the Dempster-Shafer fusion process, we aim to adaptively adjust the weight of training loss component during training loop, enhancing model performance while reducing the time needed to achieve optimal accuracy.

    \subsection{Evidence Theory and Information Fusion} 
    Significant advancements have been made in object detection with the refinement of the Intersection-over-Union (IoU) metric and the application of Dempster-Shafer (D-S) theory. These improvements have enhanced both the accuracy and reliability of detection systems across various industries. For instance, integrating IoU with cosine similarity has helped filter redundant detection boxes more efficiently, improving the precision of object detection \cite{9985205}. Techniques such as deep IoU networks, designed for dense object environments like rebar detection, and weighted deformable convolution networks combined with IoU-boundary loss for objects with irregular shapes, further demonstrate the adaptability of these methods \cite{10034986, 9797507}.
    
    Dempster-Shafer theory has proven to be an invaluable tool for managing uncertain and imprecise data, thus enhancing data reliability and detection accuracy in diverse fields \cite{6681740}. In the automotive industry, it has played a pivotal role in advancing sensor fusion, enabling more dependable object detection under uncertain conditions. Similarly, the integration of data from various sources in surveillance systems has improved the accuracy of threat detection \cite{9765125, HuoMartinez2022}. D-S theory has also been used to merge information from different imaging techniques in the medical field, leading to improved diagnostic imaging \cite{NeagoeGhenea2022}. Furthermore, combining D-S theory with neural networks has demonstrated its utility in tasks such as image classification and object recognition \cite{7528037, 65370}.
    
    The fusion of multi-modal data has greatly improved the robustness and accuracy of detection systems. This includes the integration of visual and LIDAR data for better object recognition in autonomous vehicles, as well as the combination of RGB and thermal imaging for enhanced surveillance under low-visibility conditions \cite{7314197}. Advanced neural network architectures and decision-level fusion techniques have played a crucial role in these advancements. The application of Dempster-Shafer theory in evidence-based fusion provides a robust framework for combining probabilistic model outputs, particularly in scenarios involving conflicting or incomplete data, thereby increasing detection confidence \cite{9624653, 7528037, 9765125, 4072042}. 
    
    Our method leverages this feature of D-S Theory to quantify the uncertainty of predictions relative to the ground truth in each training loop. This uncertainty is then mapped to a scoring system that adjusts the weight of the training loss for the feedback loop in the subsequent epoch. This approach enables the model to focus more on uncertain or ambiguous predictions, refining the training process for improved accuracy and robustness.

\section{Framework of Evidence Theory in Object Detection}

\begin{figure}[ht]        
        \centering
        \includegraphics[width=0.3\textwidth,keepaspectratio]{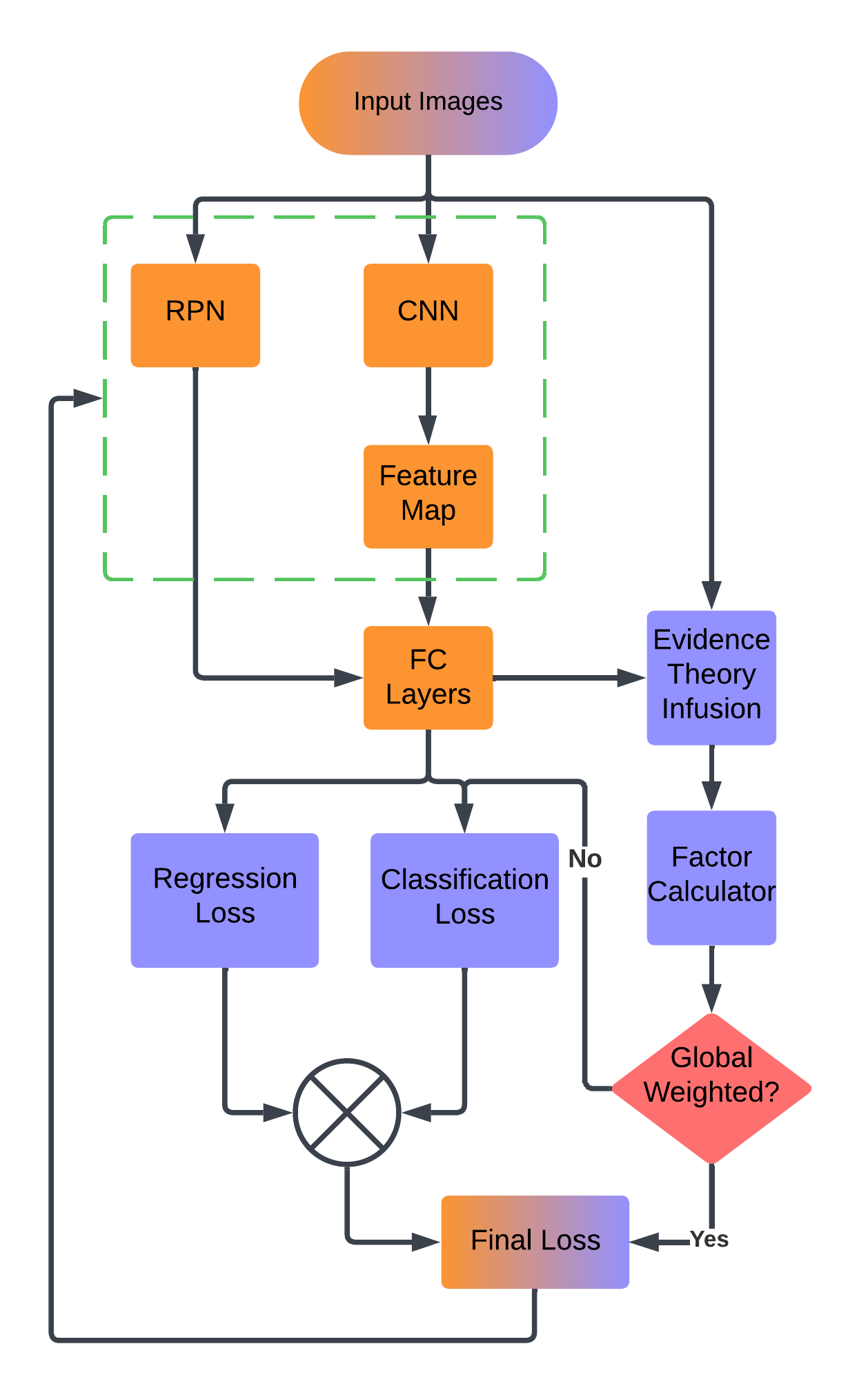}
        \caption{Training with Evidence-Enhanced Loss Function Framework}\label{fig: Training with Evidence - Enhanced Loss Function}
    \end{figure}

    The proposed framework builds upon the Faster R-CNN model, a widely used architecture for object detection tasks that can identify and classify multiple objects within an image. As illustrated in \Cref{fig: Training with Evidence - Enhanced Loss Function}, the process begins with input images fed into a Convolutional Neural Network (CNN) to extract essential features and generate a feature map. This feature map simplifies the raw image data by highlighting critical aspects needed for object recognition—such as edges, textures, and patterns—while discarding irrelevant information.
    
    The Region Proposal Network (RPN) then analyzes this feature map to suggest specific regions of interest within the image that are likely to contain objects. By predicting bounding boxes around these regions, the RPN allows the model to focus on relevant parts of the image, effectively narrowing down the search space. These proposed regions are subsequently processed through Fully Connected (FC) layers, where object classification and precise bounding box regression are performed. The FC layers refine the object predictions by determining the class labels and adjusting the position and dimensions of the bounding boxes.
    
    This initial output of object class predictions and locations serves as the input for the next phase, where an evidence-enhanced loss function is applied to further refine and enhance the model's accuracy and robustness.
    
    In the evidence-enhanced loss function, once the FC layers produce predictions for the validation images, these predictions are compared against their respective ground truth values. This comparison is processed through Evidence Theory, which builds evidence based on the level of agreement between the predictions and the ground truth. The evidence is combined to compute an uncertainty measure that reflects the confidence in the predictions. This uncertainty is then fed into a multiplication factor calculation module, where it is used to derive a weighted adjustment ratio based on a predetermined scoring card.
    
    The multiplication factor is generated using two approaches: a targeted method and a cumulative method. The targeted approach focuses solely on the uncertainty in the current data stream, generating a factor based on the specific characteristics of that stream. In contrast, the cumulative method takes into account the uncertainty from all previous data streams, averaging the results to create a more generalized factor. This factor is crucial as it controls the weighting of the feedback losses during training.
    
    There are two strategies for applying this weighted factor to the loss. In one approach, a global weighted loss is applied to both the classification loss and regression loss, allowing the factor to influence all aspects of the feedback. In the alternative approach, only the classification loss is weighted, isolating the adjustment to the object class predictions while leaving the localization unaffected. Once the weighted loss is generated, it is fed back into the training loop, allowing the model to adjust its parameters more effectively based on the reliability of its predictions. This feedback process enhances the robustness of the model by ensuring that uncertain predictions contribute less to the learning process, ultimately leading to improved accuracy and stability in object detection.
    
\section{Evidence Theory Fusion in Object Detection}
\label{ch: fusion}

The proposed algorithm leverages Evidence Theory to enhance object detection systems by effectively managing uncertainty. By consolidating varied predictions, it reduces the uncertainty of individual models, thereby increasing confidence in detection outcomes. This is particularly important in high-precision applications such as autonomous vehicle navigation and surveillance, where accurate and timely object recognition is essential.

The integration process occurs in two key stages: first, constructing Evidence Theory to assess and combine detections based on their reliability, and second, fusing this information to produce a unified detection outcome from the accumulated evidence.

\subsection{Building Evidence Theory}
Evidence Theory, also known as Dempster-Shafer Theory, offers a mathematical framework for combining evidence from different sources and making decisions based on the aggregated information. The core concepts include the mass function, belief function, and plausibility function \cite{Shafer1976,Cheng1988,Yager1987,ArevaloIbrahim2022}.

\subsubsection{Basic Probability Assignment (BPA)}\label{evidence_theory}
The mass function, also referred to as the Basic Probability Assignment (BPA), assigns a value between 0 and 1 to each subset of the frame of discernment \cite{Cheng1988}. Mathematically, it is defined as:

\begin{equation}
\label{eq: BPA}
m: 2^\Theta \rightarrow [0,1],
\end{equation}

where $\Theta$ represents the frame of discernment, and $2^\Theta$ denotes the power set of $\Theta$. The BPA must satisfy the following conditions:

\begin{itemize}
\item $m(\emptyset) = 0$,
\item $\sum_{A \subseteq \Theta} m(A) = 1$,
\end{itemize}

\subsubsection{Belief and Plausibility Functions}
The belief function (Bel) and plausibility function (Pl) are defined for all subsets of $\Theta$ and represent the degrees of belief and plausibility, respectively \cite{9624653}. These functions are expressed as:

\begin{equation}
\label{eq: Bel}
Bel(A) = \sum_{B \subseteq A} m(B),
\end{equation}

\begin{equation}
\label{eq: Pl}
Pl(A) = 1 - Bel(\overline{A}) = \sum_{B \cap A \neq \emptyset} m(B),
\end{equation}

where $\overline{A}$ represents the complement of $A$ in $\Theta$. The belief function measures the minimum support for a hypothesis, while the plausibility function quantifies the extent to which the evidence does not refute it.

\subsection{Dempster's Rule of Combination (D-S)}
Dempster's Rule of Combination is used to fuse BPAs from two independent sources of evidence \cite{Shafer1976}. Given two BPAs, $m_1$ and $m_2$, the combined BPA, denoted by $m_{1,2}$, is defined as:

\begin{equation}
\label{eq: DS}
m_{1,2}(A) = (m_1 \oplus m_2)(A),
\end{equation}
with,
\begin{equation}
\label{eq: DS2}
(m_1 \oplus m_2)(A) = \frac{1}{1-K} \sum_{B \cap C = A} m_1(B) m_2(C) ,
\end{equation}

\begin{equation}
\label{eq: DS-K}
K = \sum_{B \cap C = \emptyset} m_1(B) m_2(C),
\end{equation}

where $K$ is a normalization constant that accounts for the conflict between the two sources of evidence \cite{7528037}.

These mathematical principles make Dempster-Shafer Theory a versatile and powerful tool for integrating evidence and managing uncertainty across a wide range of decision-making applications \cite{65370}.
In our study, the analytical framework focuses on a specific group, termed the frame of discernment, represented as $\Theta = \{m_{x_1}, m_{x_2}, ..., m_{x_N} \}$, where $N$ is the total number of labels in the group and is a natural number ($N \in \mathbb{N}$). This group includes each subset characterized by a prediction label, bounding box, and associated score. When a group contains multiple prediction labels, Evidence Theory is applied to conduct a unified analysis of these diverse predictions.

The predicted label and its corresponding score are mapped to a dictionary variable, where $E_{x} = \{L_{x}, S_{x}\}$.Here, $L_{x}$ refers to the label, and $S_{x}$ is the predicted confidence score. This structured approach facilitates a more accurate and reliable analysis of the prediction labels within the group using Evidence Theory principles. 


As discussed in \Cref{evidence_theory}, Basic Probability Assignments (BPAs) must sum to one, as expressed in the equation $\sum_{A \subseteq \Theta} m(A) = 1$. Recognizing that the total BPA within a group may not always equal one, we propose a method for converting predictions into mass functions. This approach deviates from earlier strategies, such as those proposed by Arevalo et al. \cite{ArevaloPiolo2022}, and introduces a normalization process for the frame of discernment.

Our method assumes that each prediction is a true reflection of an object, potentially assigning Basic Probability Assignment (BPA) values exceeding one. To address this, we normalize the frame of discernment, yielding a revised frame: 
\begin{equation}
\label{eq: hat_thetha}
\hat\Theta = \sum_{x=1}^N \hat m_{x} = \{\hat m_{x_1}, \hat m_{x_2}, \dots, \hat m_{x_N} \},
\end{equation}
where masses are adjusted to ensure that the sum of probabilities across all subsets of the frame equals one, aligning with the core principles of BPA. Each normalized mass \(\hat m(x)\) is calculated as \(\hat m(x) = \hat E_{x} = \{L_{x}, \xi_{x}\}\). The normalization process divides each individual score by the sum of all scores for that group’s masses, as shown in Equation \ref{eq: hat_xi}, ensuring that the total mass aligns with the D-S Theory constraint of \(\sum_{A \subseteq \Theta} \hat m(A) = 1\). The mass for each prediction label \(x\) is given by:
\begin{equation}
\label{eq: hat_xi}
\hat \xi_{x} = \frac{S_{x}}{\sum\limits_{x=N = \emptyset} S_{x_N}},
\end{equation}
where, \(\hat \xi_{x}\) represents the normalized score for each predicted label, ensuring that the total mass across all predictions equals one. This normalization process not only maintains consistency with D-S Theory but also optimizes the interpretability and accuracy of the model’s predictions in uncertain scenarios.

\subsection{Fusion of Classification Information}
\label{sub.ch: clf_fusion}

Once the masses have been normalized, we construct an evidence theory matrix of dimensions $N \times N$, denoted by $\textbf{H}$:
\begin{equation}
\label{eq: ET_matrix}
\resizebox{0.8\columnwidth}{!}{%
$\textbf{H} = 
\renewcommand{\arraystretch}{1.5} 
\begin{array}{c|c|ccc|}
                                & \hat E_{y_1}/\hat L_{y_1}     & \hat E_{y_2}/\hat L_{y_2}     & \cdots    & \hat E_{y_N}/\hat L_{y_N} \\
    \cline{1-5}
    \hat E_{x_1}/\hat L_{x_1}   & \xi_{x_1} \cdot \xi_{y_1}     & \xi_{x_1} \cdot \xi_{y_2}     & \cdots    & \xi_{x_1} \cdot \xi_{y_N}  \\
    \cline{1-5}
    \hat E_{x_2}/\hat L_{x_2}   & \xi_{x_2} \cdot \xi_{y_1}     & \xi_{x_2} \cdot \xi_{y_2}     & \cdots    & \xi_{x_2} \cdot \xi_{y_N} \\
    \vdots                      & \vdots                        & \vdots                        & \ddots    & \vdots \\
    \hat E_{x_N}/\hat L_{x_N}   & \xi_{x_N} \cdot \xi_{y_1}     & \xi_{x_N} \cdot \xi_{y_2}     & \cdots    & \xi_{x_N} \cdot \xi_{y_N} \\
    \cline{1-5}
\end{array}$
},
\end{equation}
where, the matrix captures all possible permutations of the scores associated with each mass. Each element of the matrix, $\textbf{H}_{ij}$, is defined as the product of the scores of two distinct masses, represented by $\hat E_{x_i}$ and $\hat E_{x_j}$. Given that each mass is tagged with a specific label, we can reinterpret this matrix through the lens of prediction labels. The interactions between any two labels are quantitatively expressed by the product of their normalized scores. This dual-layer representation enriches the matrix by offering insights into both the quantitative strength of predictions and the qualitative relationships between labels. This enhances the utility of the matrix in synthesizing information from diverse predictions and facilitates a deeper understanding of the data.

This matrix plays a crucial role in assessing the likelihood of events under certain conditions, as guided by Dempster-Shafer (D-S) theory. According to the D-S rule, as outlined in \Cref{eq: DS_2m}, evidences (masses) can be combined in pairs using the formula:
\begin{equation}
\label{eq: DS_2m}
m_{1,2}(A) = (m_1 \oplus m_2)(A),
\end{equation}
whereas, to handle multiple masses, we employ a sequential fusion strategy, applying the D-S rule iteratively by combining two masses at a time until all are merged into a single probabilistic outcome. To combine multiple masses, \Cref{eq: DS_2m} can be extended to:
\begin{equation}
\label{eq: DS_3m}
m_{1,2,3}(A) = ((m_{1,2}) \oplus m_3)(A) = ((m_1 \oplus m_2) \oplus m_3)(A),
\end{equation}
This enables the efficient synthesis of diverse pieces of evidence into a unified probability assessment, enhancing decision-making with a more comprehensive understanding of event probabilities. Furthermore, \Cref{eq: DS_2m} can be extended to an infinite number of fusions, though this can become computationally intensive. The general approach for fusing an \( N \) number of masses is given by:
\begin{equation}
\label{eq: DS_seq}
m_{1,2,3,\dots,N}(A) = (m_{1,2,\dots,N-1} \oplus m_N)(A),
\end{equation}
where each pair of sequential masses is fused together first, and then the resulting product is fused with the next mass.

Using the matrix from \Cref{eq: ET_matrix}, we derive fusion outcomes for all combinations of predicted labels across pairs of masses. The diagonal elements of the matrix correspond to the combined Basic Probability Assignment (BPA) as shown in \Cref{eq: DS2}, representing the certainty of each prediction after fusion. The off-diagonal elements contribute to the calculation of the uncertainty factor $K$, as specified in \Cref{eq: DS-K}. The overall certainty of the fusion, denoted as $\Phi$, is determined by subtracting the uncertainty $K$ from unity, i.e., $\Phi = 1 - K$. This process allows for a nuanced assessment of both certainty and uncertainty, enabling a comprehensive analysis of the predictive outcomes.

\begin{equation}
\label{eq: phi}
\Phi = 1-\sum_{B \cap C = \emptyset} m_1(B) m_2(C) = 1-K,
\end{equation}

\section{Uncertainty Based Training Models}
\subsection{Building Evidence Theory and Extracting Uncertainty}

In object detection models, Evidence Theory (ET) is constructed using both the predicted labels and the ground truth. These two sources of information are fused using the Dempster-Shafer rule of combination, allowing us to capture the uncertainty associated with predictions. This subsection outlines how predictions and ground truth are used to build Evidence Theory and how uncertainty is extracted through their combination.

\subsubsection{Prediction Labels, Ground Truth, and Softmax Scores}

For each detected object, the model generates a set of predicted labels, along with confidence scores produced by the softmax function, which represent the likelihood of each label being correct. Simultaneously, the ground truth provides the actual labels, allowing us to compare predictions against known outcomes.

The prediction labels and their associated softmax scores are treated as evidence from the model, while the ground truth serves as an additional source of evidence. In Evidence Theory, both the predictions and ground truth are treated as mass functions. The mass function \(m_1\) for predictions is defined as:
\begin{equation}
m_1(\text{label}_i) = \text{softmax}(\text{score}_i),
\end{equation}
The mass function \(m_2\) can originate from various heterogeneous sources, including experts who manually annotate the data, predefined ground truth datasets, or other methods such as automated labeling systems. When the source provides the predefined ground truth datasets, it assigns full belief to the correct label. This is mathematically represented as:
\begin{equation}
m_2(\text{ground truth}) \simeq 1,
\end{equation}
This equation indicates that all the belief mass is assigned to the ground truth label, reflecting complete certainty in its correctness. There is no ambiguity or uncertainty, and no belief is allocated to any other possible labels. The mass function for the unknown hypothesis, \(\theta\), is assigned as:
\begin{equation}
m_1(\theta) = 1 - \sum m_1(\text{label}_i),
\end{equation}

\subsubsection{Fusing Evidence with Dempster-Shafer Combination}

To effectively capture the uncertainty, the Dempster-Shafer rule of combination is applied to fuse the evidence from the predictions and the ground truth. The combination rule is formulated as follows:
\begin{equation}
m(\text{label}) = \frac{\sum m_1(\text{label}_i) \cdot m_2(\text{label}_i)}{1 - \sum m_1(\text{label}_i) \cdot m_2(\text{label}_i)},
\end{equation}
where the denominator normalizes the combined belief, accounting for conflicting evidence between the predictions and ground truth.

This combination process merges the model’s predictions with the certainty provided by the ground truth, effectively integrating multiple pieces of evidence into a single mass function. The resulting mass function represents the refined belief for each label, incorporating both sources of information.

\subsubsection{Extracting Uncertainty}

Uncertainty is extracted by analyzing the resulting mass function after the combination. When the mass is concentrated on a single label, the uncertainty is low, indicating a high level of confidence. On the other hand, when the mass is distributed among multiple labels or heavily weighted toward the unknown hypothesis \(\theta\), the uncertainty is high.

The total uncertainty is quantified using the belief (\(Bel\)) and plausibility (\(Pl\)) values associated with the unknown hypothesis:
\begin{equation}
K = Pl(\theta) - Bel(\theta),
\end{equation}
A higher value of \(K\) indicates greater uncertainty in the model's prediction.

Incorporating this uncertainty into the training loop enables dynamic adjustment of the feedback loss, allowing the model to focus on uncertain predictions and learn from them more effectively. This approach accelerates convergence by guiding the model's attention to regions of higher uncertainty.

\subsection{Dynamic Weighting of Uncertainty Using Scorecards}

Effectively managing uncertainty in object detection models is crucial for enhancing both performance and reliability. This section introduces a method for dynamically weighting uncertainty by employing a scorecard approach. After applying the Dempster-Shafer theory to fuse classification evidence, we obtain an uncertainty value \( K \) for each detection instance. This value represents the degree of conflict among evidence sources, with higher values indicating greater uncertainty.

To adjust the learning process based on this uncertainty, we define a multiplication factor \( w \) that scales the feedback losses during training. The factor \( w \) is determined by mapping the uncertainty value \( K \) to predefined ranges specified in a scorecard. Two different scorecards, Score Card A and Score Card B, are utilized in this research, as shown in  \Cref{eq:score-card-A} and \Cref{eq:score-card-B}, respectively.

The scorecards act as hyperparameters for the model, allowing for fine-tuning based on the dataset and feature characteristics. By adjusting the multiplication factors associated with different uncertainty ranges, we can influence the model's focus during training, emphasizing or de-emphasizing certain instances according to their uncertainty.

The mapping from uncertainty values to multiplication factors can be formalized as follows. Let \( K \) be the uncertainty value obtained from the Dempster-Shafer combination, where \( K \in [0, 100] \). Define a piecewise function \( w(K) \) that assigns a multiplication factor based on the range in which \( K \) falls:

\paragraph{For Score Card A:}

\begin{equation}
\label{eq:score-card-A}
w(K) =
\begin{cases}
2.3, & \text{if } 0.80 < K \leq 1.00 \\
2.0, & \text{if } 0.60 < K \leq 0.80 \\
1.7, & \text{if } 0.40 < K \leq 0.60 \\
1.4, & \text{if } 0.30 < K \leq 0.40 \\
1.1, & \text{if } 0.20 < K \leq 0.30 \\
0.8, & \text{if } 0.10 < K \leq 0.20 \\
0.5, & \text{if } 0.00 \leq K \leq 0.10 \\
\end{cases}
\end{equation}

\paragraph{For Score Card B:}

\begin{equation}
\label{eq:score-card-B}
w(K) =
\begin{cases}
1.5, & \text{if } 0.80 < K \leq 1.00 \\
1.1, & \text{if } 0.60 < K \leq 0.80 \\
0.8, & \text{if } 0.40 < K \leq 0.60 \\
0.5, & \text{if } 0.20 < K \leq 0.40 \\
0.2, & \text{if } 0.00 \leq K \leq 0.20 \\
\end{cases}
\end{equation}

These functions assign multiplication factors \( w(K) \) based on the uncertainty value \( K \), effectively scaling the loss function for each training instance. The higher the uncertainty, the greater the multiplication factor in Score Card A, indicating that instances with higher uncertainty are given more weight during training. In contrast, Score Card B assigns lower multiplication factors to higher uncertainty ranges, potentially reducing the emphasis on uncertain instances.

During the training process, the multiplication factor \( w(K) \) is used to adjust the loss function \( L \) for each instance. The adjusted loss \( L' \) is computed as:

\begin{equation}
L' = w(K) \times L.
\end{equation}

By incorporating \( w(K) \) into the loss calculation, the model dynamically adjusts its learning based on the uncertainty associated with each instance. This approach allows for a more nuanced training process, potentially improving overall model performance by appropriately balancing the influence of uncertain instances.

By treating the scorecards as hyperparameters, we can experiment with different weighting schemes to find the optimal balance for a given dataset and set of features. The multiplication factors and uncertainty ranges can be adjusted based on validation performance, allowing for customized training strategies that suit specific applications.

The dynamic weighting of uncertainty using scorecards provides a flexible and effective means of incorporating uncertainty into the training of object detection models. By adjusting the loss function based on uncertainty values, the model can be guided to focus on instances that are most beneficial for learning, potentially improving performance and generalization.

\subsection{Injection of Uncertainty into the Feedback Loss}

This experimental design introduces a methodology for integrating uncertainty estimation into the training process of object detection models and evaluates its impact on performance. The framework enhances the Faster R-CNN model by incorporating an improved loss calculation method that includes uncertainty-based multiplication factors. By refining the loss function through an evidence-based fusion approach, the framework aims to improve the model's performance metrics.

\begin{figure}[h]
    \centering
    \includegraphics[width=0.5\textwidth,keepaspectratio]{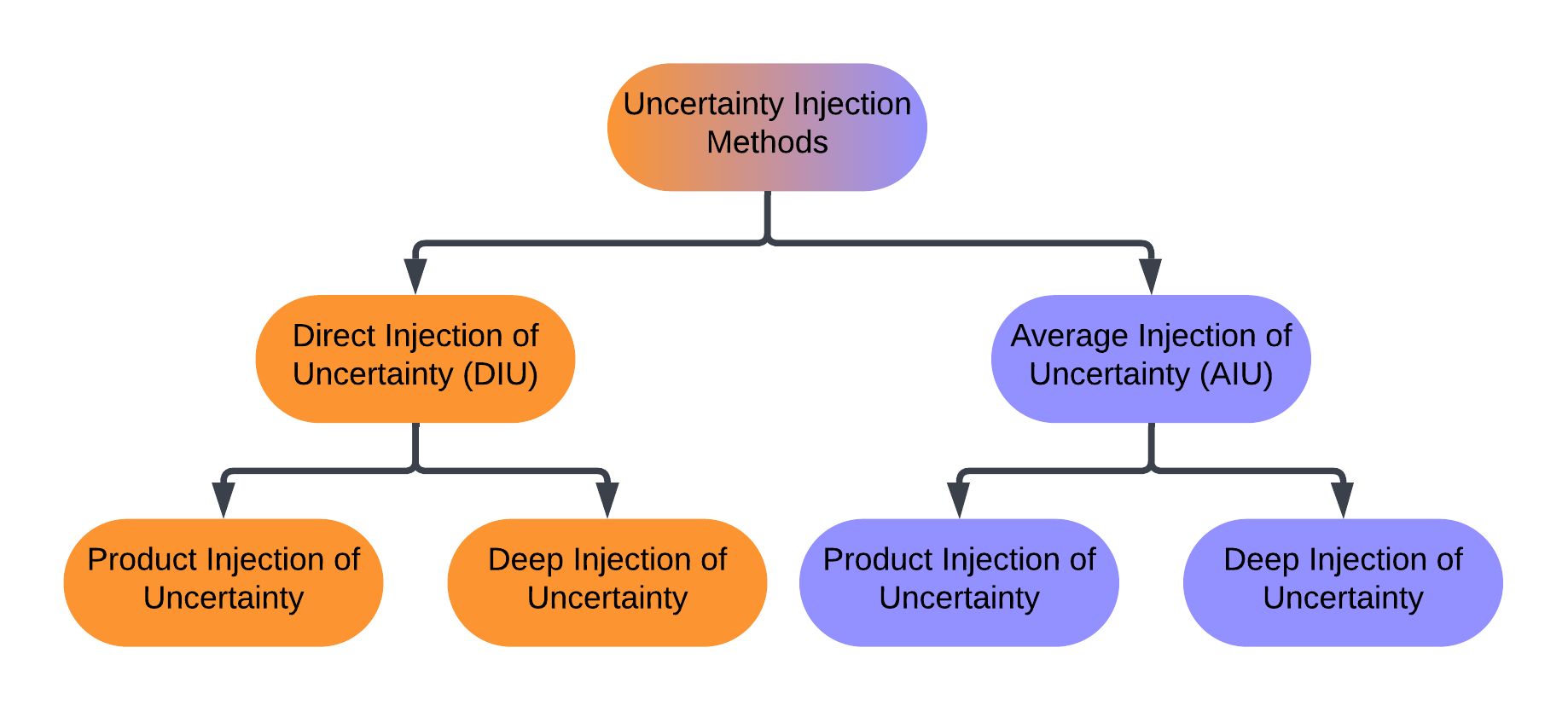}
    \caption{Hierarchical Structure of Uncertainty Injection Methods}
    \label{fig: Tree structure for Uncertainty Injection Methods}
\end{figure}

In this framework, the uncertainty injection process is structured around two main dimensions: how to inject uncertainty and where to inject it. The two primary methods, Direct Injection of Uncertainty (DIU) and Average Injection of Uncertainty (AIU), define the "how to inject" by specifying the approach for incorporating uncertainty into the model. DIU introduces uncertainty factors directly and individually into each prediction, enhancing sensitivity to real-time input variations, while AIU takes an averaged approach, aggregating uncertainty over multiple predictions or time steps, which smooths out potential fluctuations. The "where to inject it" is determined by the two sub-approaches: Product Injection of Uncertainty and Deep Injection of Uncertainty. Product Injection introduces uncertainty factors as multiplicative elements directly into the total loss function, focusing on individual predictions’ confidence. In contrast, Deep Injection integrates uncertainty within classification loss only, allowing for a more hierarchical and comprehensive handling of uncertainty that accumulates across multiple model stages. \Cref{fig: Tree structure for Uncertainty Injection Methods} presents the hierarchical structure of these uncertainty injection methods, showing how the two dimensions interact within the framework.

\subsubsection{Direct Injection of Uncertainty (DIU)}
\label{subsec: Direct Injection of Uncertainty}

\begin{figure}[h]
    \centering
    \includegraphics[width=0.2\textwidth,keepaspectratio]{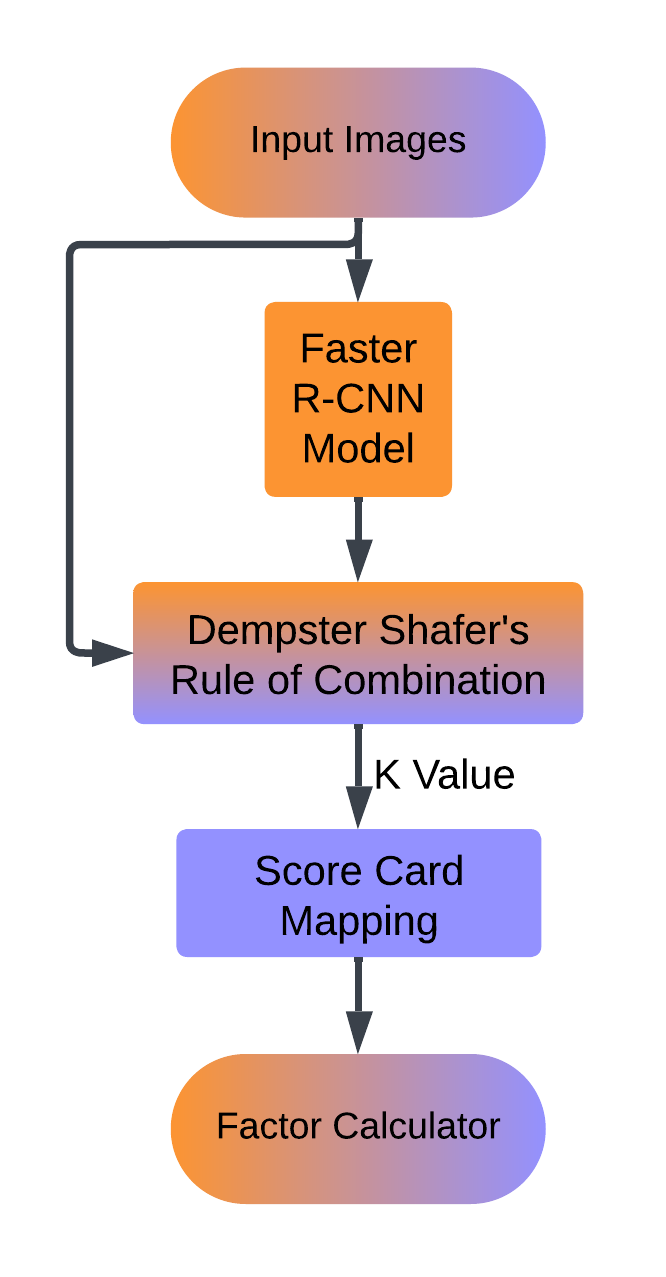}
    \caption{Framework of Direct Injection of Uncertainty}
    \label{fig: DIU}
\end{figure}

The Direct Injection of Uncertainty (DIU) framework integrates uncertainty estimation directly into the object detection process to enhance model performance. This method employs the Dempster-Shafer theory of evidence to quantify and manage uncertainties, leading to a more adaptive loss function. \Cref{fig: DIU} illustrates the structure of Direct Injection of Uncertainty (DIU) framework in the system.

The key components of the DIU framework are:

\begin{enumerate}
    \item[1.] \textbf{Validation Images and Predicted Values:} A set of validation images is used to evaluate the model's performance. For each image, the model generates predicted values, including object classes and bounding box coordinates. These predictions are compared with ground truth values to assess accuracy.
    \item[2a.] \textbf{Dempster-Shafer Rule of Combination}: The Dempster-Shafer theory is employed to quantify uncertainty by combining the predictions from multiple sources with the ground truth values, yielding a measure of uncertainty, denoted as the $K$ value. In this framework, the combination of Basic Probability Assignments (BPAs) from two independent sources, $m_1$ and $m_2$, produces a combined BPA, $m_{1,2}$, as follows:
    \begin{equation} 
        m_{1,2}(A) = (m_1 \oplus m_2)(A), 
    \end{equation}    
    where the combined mass $m_{1,2}(A)$ is calculated by:    
    \begin{equation} 
        (m_1 \oplus m_2)(A) = \frac{1}{1-K} \sum_{B \cap C = A} m_1(B) m_2(C), 
    \end{equation}    
    with the conflict factor $K$ defined as:    
    \begin{equation} 
        K = \sum_{B \cap C = \emptyset} m_1(B) m_2(C). 
    \end{equation}    
    Here, $K$ serves as a normalization constant that accounts for conflicting evidence between the two sources. This approach, grounded in Dempster-Shafer theory, enables the model to quantify and manage uncertainties more effectively, enhancing the robustness of predictions in uncertain conditions.
    
    \item[3.] A predefined scorecard maps the uncertainty measure, $K$, to a corresponding multiplication factor $w(K)$. This factor is then applied to adjust the feedback loss during training, ensuring that the model is responsive to varying levels of certainty in its predictions. Specifically, this multiplication factor is integrated into the loss function, modifying the loss based on the degree of uncertainty. The process is formalized through the following equation:    
    \begin{equation} 
        L' = w(K) \times L, 
    \end{equation}   
    where $w(K)$ is a function that defines the relationship between the $K$ value and the multiplication factor, as determined by the scorecard mentioned in \Cref{eq:score-card-A} and \Cref{eq:score-card-B}. This adaptive loss adjustment enables the model to emphasize more confident predictions, thereby refining the learning process and contributing to improved model accuracy and stability.
\end{enumerate}

\subsubsection{Average Injection of Uncertainty (AIU)}
\label{subsec: Average Injection of Uncertainty}

\begin{figure}[h]
    \centering
    \includegraphics[width=0.25\textwidth,keepaspectratio]{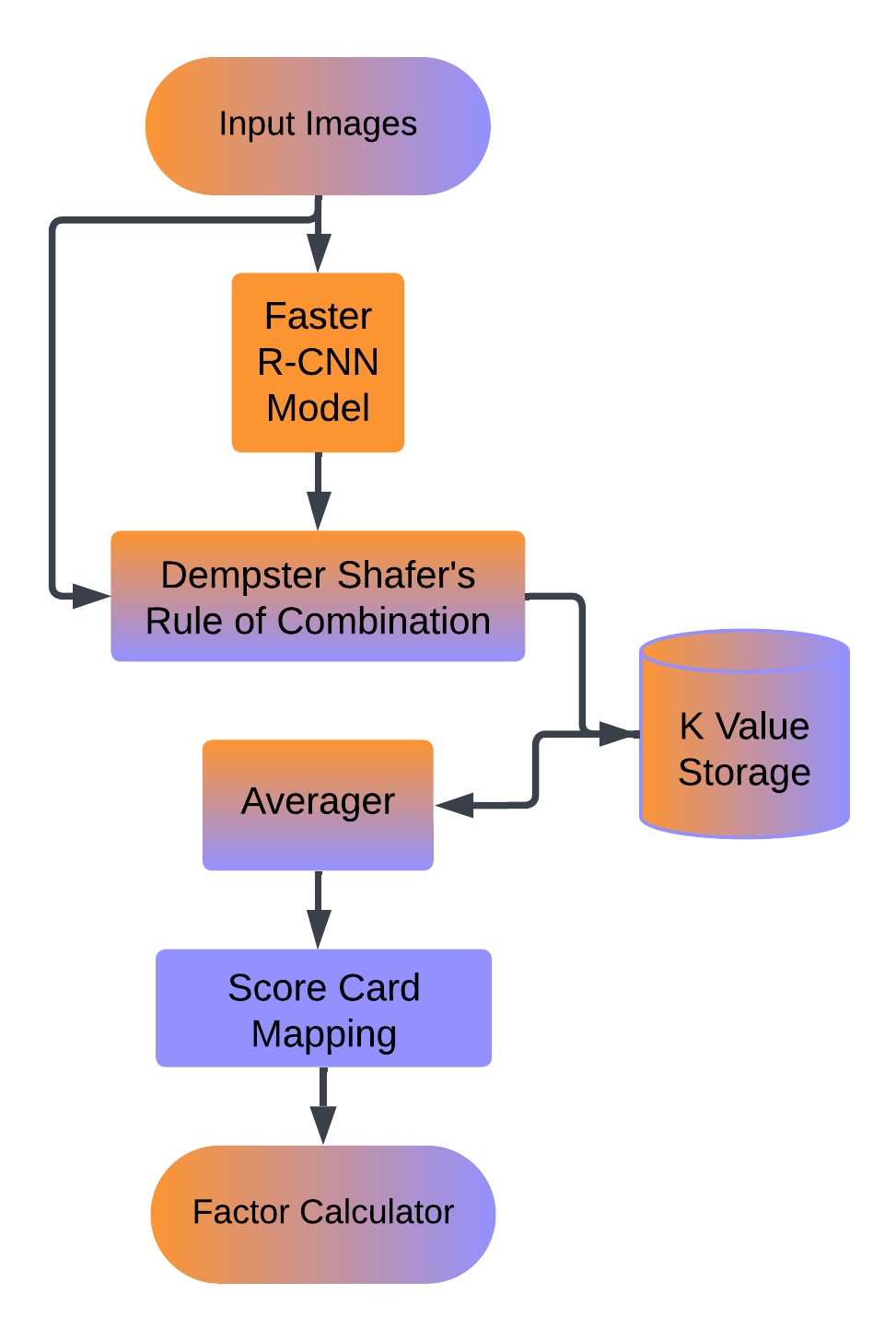}
    \caption{Framework of Average Injection of Uncertainty}
    \label{fig: AIU}
\end{figure}

The Average Injection of Uncertainty (AIU) framework builds upon DIU by incorporating an averaging mechanism for $K$ values across validation cycles. This approach aims to maintain consistent uncertainty management during training. \Cref{fig: AIU} illustrates the structure of Direct Injection of Uncertainty (DIU) framework in the system.

Key aspects of the AIU framework include:

\begin{enumerate}
    \item[2b.] \textbf{Averaging of Current and Previous \( K \) Values:} To ensure consistency in uncertainty assessment over time, the \( K \) values obtained from the current validation cycle are averaged with the \( K \) values from previous epochs. This averaged \( K \) value, denoted as \( K' \), provides a stabilized measure of uncertainty, which is then used to map to the appropriate multiplication factor in the scorecard. By averaging \( K \) values across epochs, the model adapts more gradually to changes in uncertainty, resulting in a smoother training process and more reliable adjustment of the feedback loss. The averaging process is formalized as follows:

    \begin{equation}
        \label{eq:K-prime}
        K' = \frac{1}{n} \sum_{i=1}^{n} K_i,
    \end{equation}
    
    where \( K' \) is the averaged uncertainty measure, \( K_i \) represents the \( K \) values from each of the previous \( n \) epochs, and \( n \) is the number of epochs considered. This averaged \( K' \) value is then used to determine the multiplication factor from the scorecard mentioned in \Cref{eq:score-card-A} and \Cref{eq:score-card-B}, allowing for a consistent and adaptive training process based on smoothed uncertainty levels.

\end{enumerate}

\subsubsection{Product Injection of Uncertainty (PIU)}

\begin{figure}[h]
    \centering
    \includegraphics[width=0.3\textwidth,keepaspectratio]{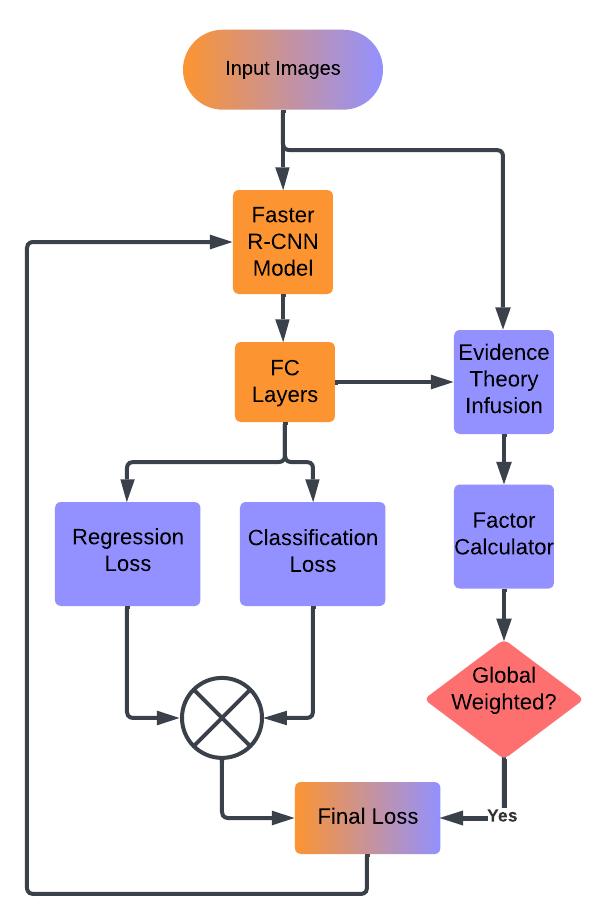}
    \caption{Product Injection of Uncertainty}
    \label{fig: Multiplication Approach}
\end{figure}

In this approach, the total loss (the sum of classification and localization losses) is scaled using the multiplication factor $w(K)$, derived from either the DIU or AIU frameworks, as illustrated in \Cref{fig: Multiplication Approach}.

\subsubsection{Deep Injection of Uncertainty }

\begin{figure}[h]
    \centering
    \includegraphics[width=0.3\textwidth,keepaspectratio]{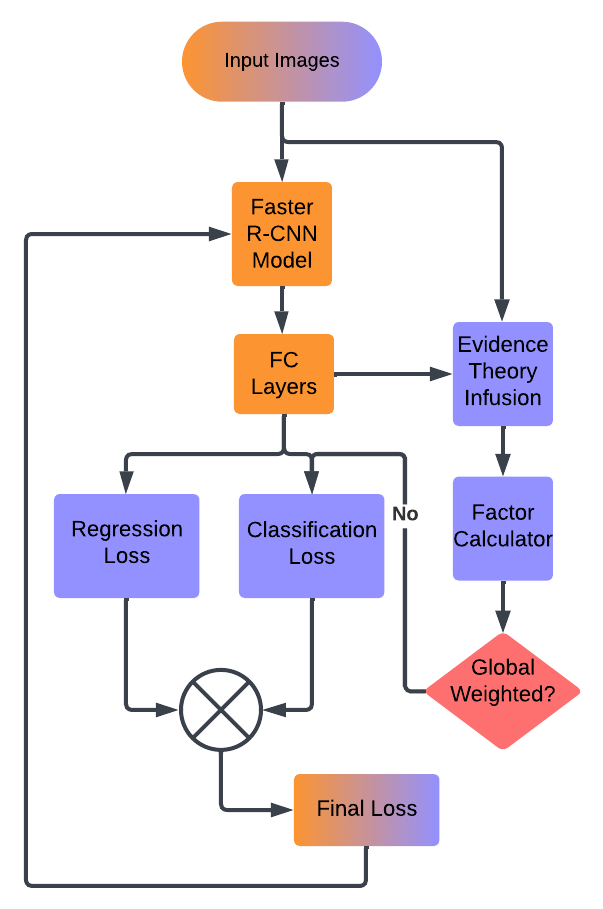}
    \caption{Deep Injection of Uncertainty}
    \label{fig: Deep Multiplication Approach}
\end{figure}

In contrast to the Product Injection approach, the Deep Injection of Uncertainty framework scales only the classification loss using the multiplication factor $w(K)$, while the localization loss remains unchanged. This method is depicted in \Cref{fig: Deep Multiplication Approach}.

\subsection{Performance Metrics}

To evaluate the novel training approach, we introduce the Performance Score, calculated as:

\begin{equation} 
    S_{Performance} = mAP - (L_{Training} + L_{Validation})
\end{equation}

This metric provides a holistic assessment by combining Mean Average Precision (mAP) with cumulative training and validation losses. Unlike traditional metrics that focus solely on output accuracy, the Performance Score accounts for both precision and training efficiency. It emphasizes models that achieve high accuracy through effective learning, penalizing those that do not train efficiently. Higher Performance Scores indicate better overall model performance, offering a more comprehensive evaluation in the field of computer vision.

\section{Baseline}

\subsection{Dataset (Pascal Visual Object Classes 2012)}

    The Pascal Visual Object Classes (VOC) 2012 dataset \cite{pascal-voc-2012} is a cornerstone in computer vision research, extensively used for benchmarking object detection and image classification algorithms. It comprises a rich collection of images annotated with object instances and their corresponding bounding boxes across 20 diverse categories, including animals, vehicles, and household items. The dataset is acclaimed for its diversity in object scales, poses, lighting conditions, and occlusions, which present substantial challenges and drive the development of robust and generalizable computer vision models.
    
    The dataset is typically divided into standard splits of training, validation, and test sets, ensuring effective training and evaluation of models. \Cref{table:voc2012_classes} lists all the object classes included in the Pascal VOC 2012 dataset along with the counts of annotated instances in the training, validation, and test sets. This detailed breakdown highlights the distribution of objects within each split of the dataset, with the "Person" class having the highest frequency across all splits, reflecting its prevalence in real-world scenarios.
    
    \begin{table}[h]
        \centering
        \caption{Pascal VOC 2012 Object Classes and Instance Counts in Training, Validation, and Test Sets}
        \label{table:voc2012_classes}
        \begin{tabular}{|l|c|c|c|}
            \hline
            \textbf{Class} & \textbf{Training} & \textbf{Validation} & \textbf{Testing} \\
            \hline
            Aeroplane & 532 & 214 & 256 \\
            Bicycle & 496 & 169 & 172 \\
            Bird & 779 & 210 & 282 \\
            Boat & 608 & 181 & 270 \\
            Bottle & 895 & 280 & 386 \\
            Bus & 378 & 122 & 185 \\
            Car & 1,403 & 471 & 618 \\
            Cat & 674 & 258 & 345 \\
            Chair & 1,779 & 517 & 760 \\
            Cow & 442 & 127 & 202 \\
            Dining Table & 445 & 122 & 233 \\
            Dog & 923 & 300 & 375 \\
            Horse & 458 & 180 & 165 \\
            Motorbike & 454 & 145 & 202 \\
            Person & 9,601 & 3,502 & 4,298 \\
            Potted Plant & 667 & 225 & 310 \\
            Sheep & 550 & 240 & 294 \\
            Sofa & 515 & 143 & 183 \\
            Train & 362 & 146 & 196 \\
            TV/Monitor & 518 & 180 & 195 \\
            \hline
        \end{tabular}
    \end{table}

\subsection{Framework}

    Faster R-CNN is an advanced deep learning framework for high-accuracy object detection. It processes input images through a Convolutional Neural Network (CNN) to extract feature maps shared between the Region Proposal Network (RPN) and the detection network, enhancing computational efficiency. The RPN generates object proposals with objectness scores, and these regions of interest (RoIs) are pooled to a fixed size via RoI Pooling. Fully connected layers then perform classification and bounding box regression using a multi-task loss function that combines classification and regression losses.
    
    Training involves backpropagating the combined loss to update the weights of the CNN, RPN, and fully connected layers, improving predictions over iterations. This end-to-end architecture enables efficient and accurate object detection, making it effective for real-time applications.
    
    The model uses the Adagrad optimizer with a learning rate of 0.001, adapting learning rates based on feature frequency, effective for handling sparse and varied data in object detection tasks. Pre-trained weights from ImageNet are utilized for each backbone model (VGG16, ResNet-18, EfficientNet-B0, MobileNet-v2) to enhance learning efficacy. An identical configuration of backbone models and hyperparameters is maintained throughout this research.

\subsection{Binary Class}
This section describes the traditional training approach used to evaluate the performance of the Faster R-CNN model with various backbones. The model was trained using a single class, "chair," to assess its effectiveness. The performance scores achieved through this method are documented and will serve as a benchmark for comparing the results of new training techniques.

\Cref{tab: Performance Scores- Traditional Approach - chair} presents the highest performance scores achieved by four different models during their training. Each score represents the model's best performance at a specific epoch, indicating the point where it achieved an optimal balance between precision and loss for the "chair" class.

\captionsetup{skip=10pt} 
\begin{table}[H]
    \centering
    \caption{Model Performance Scores - Traditional Approach (Single Class: Chair)}
    \label{tab: Performance Scores- Traditional Approach - chair}
    \begin{tabular}{|c|c|c|}
        \hline
        \textbf{Model} & \textbf{Epoch} & \textbf{Performance Score} \\
        \hline
        ResNet & 13 & -0.47268 \\
        \hline
        MobileNet & 35 & -0.49351 \\
        \hline
        EfficientNet & 38 & -0.36971 \\
        \hline
        VGG & 24 & -0.58170 \\
        \hline
    \end{tabular}
\end{table}

\subsection{Multi-Class}
Building on the single-class experiment, where the Faster R-CNN model was trained using only the "chair" class, this section extends the scope to evaluate the model's performance when trained on multiple classes. Specifically, the model was trained on four distinct object classes: "bus," "car," "motorbike," and "train." This multi-class training allows us to assess the model's scalability and robustness across a broader range of object categories.

\Cref{tab: Performance Scores- Traditional Approach - 4 class} presents the highest performance scores achieved by four different models during their training on the multi-class dataset. Similar to the single-class experiment, each score represents the model's best performance at a specific epoch, indicating where it achieved the best balance between precision and loss for the combined object classes.

\captionsetup{skip=10pt} 
\begin{table}[H]
    \centering
    \caption{Model Performance Scores - Traditional Approach (Multi-Class: Bus, Car, Motorbike, Train)}
    \label{tab: Performance Scores- Traditional Approach - 4 class}
    \begin{tabular}{|c|c|c|}
        \hline
        \textbf{Model} & \textbf{Epoch} & \textbf{Performance Score} \\
        \hline
        ResNet & 16 & -0.09887 \\
        \hline
        MobileNet & 36 & -0.09674 \\
        \hline
        EfficientNet & 37 & 0.02308 \\
        \hline
        VGG & 28 & -0.26662 \\
        \hline
    \end{tabular}
\end{table}

\section{Experiments}
    \subsection{Score Card A - Binary Class}
        \subsubsection{Product Injection of Uncertainty}
             The Product Injection of Uncertainty optimizes the model’s training process by modulating the total loss through a Multiplication Factor $w(K)$, now derived from Scorecard A. As outlined in \Cref{subsec: Direct Injection of Uncertainty} and \Cref{subsec: Average Injection of Uncertainty}, distinct methodologies are employed to compute $M$, leading to varied outcomes in the model’s Performance Score.

            The Performance Scores associated with the Product Injection of Uncertainty, based on the source of the Multiplication Factor—namely, Direct Injection of Uncertainty (DIU) and Average Injection of Uncertainty (AIU)—are presented in the accompanying \Cref{tab: Model Performance Scores for DIU and AIU- Multiplication Approach- 1SC}.
            \captionsetup{skip=10pt}
            \begin{table}[H] 
                \centering
                \caption{Model Performance Scores with DIU and AIU Methods Using Product Injection of Uncertainty}
                \label{tab: Model Performance Scores for DIU and AIU- Multiplication Approach- 1SC}
                \resizebox{\columnwidth}{!}{%
                \begin{tabular}{|c|c|c|c|c|c|c|}
                    \hline
                    \textbf{Model} & \textbf{Epoch (DIU)} & \textbf{DIU} & \textbf{Epoch (AIU)} & \textbf{AIU} & \textbf{Epoch(Base.)} & \textbf{Baseline} \\
                    \hline
                    ResNet & \textbf{09} & \textbf{-0.40599} & \textbf{09} & \textbf{-0.40529} & 13 & -0.47268 \\
                    \hline
                    MobileNet & \textbf{32} & \textbf{-0.40616} & \textbf{34} & \textbf{-0.40728} & 35 & -0.49351 \\
                    \hline
                    EfficientNet & 39 & \textbf{-0.32708} & \textbf{36} & \textbf{-0.31576} & 38 & -0.36971 \\
                    \hline
                    VGG & \textbf{11} & \textbf{-0.45602} & \textbf{14} & \textbf{-0.46331} & 24 & -0.58170 \\
                    \hline
                \end{tabular}%
                }
            \end{table}

            \begin{figure}[h]
                \centering
                \includegraphics[width=0.45\textwidth,keepaspectratio]{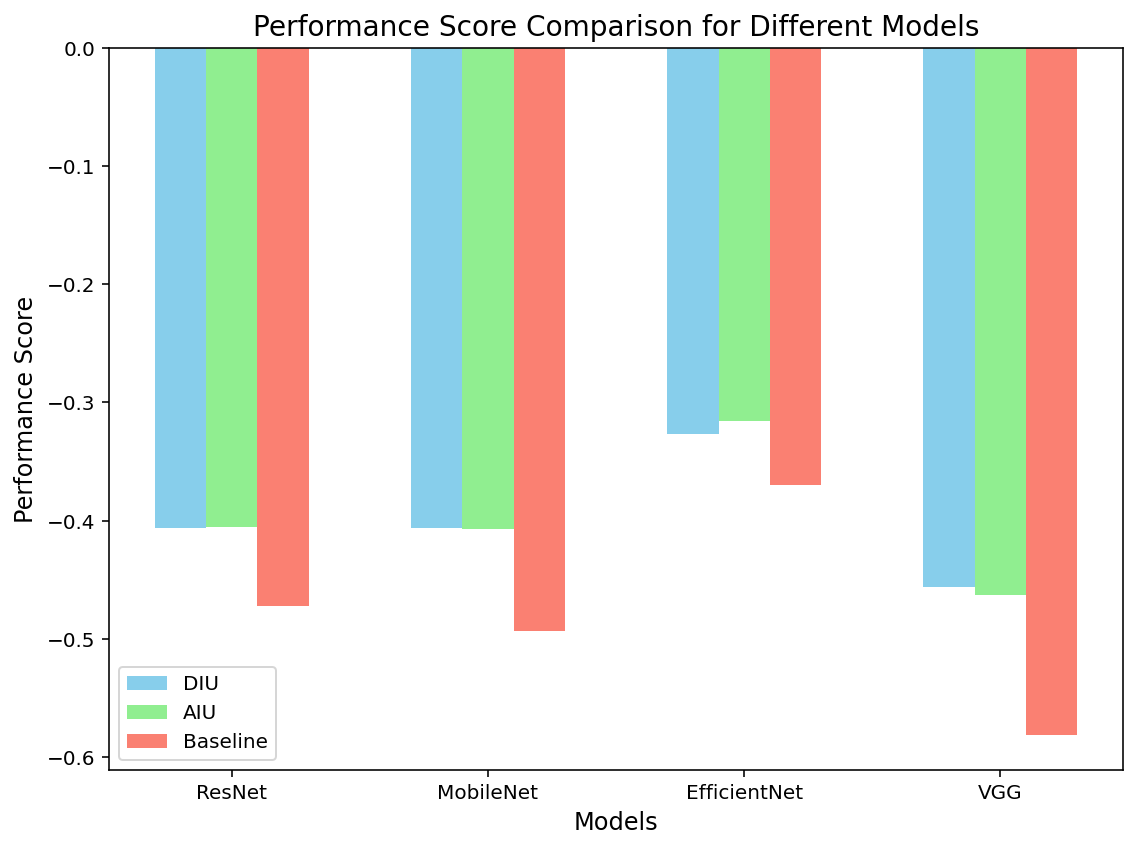}
                \caption{Product Injection of Uncertainty - Evaluation of Model Performance Across DIU, AIU, and Baseline Approaches using Score Card A}
                \label{fig:binary_class_product_injection_SC1.png}
            \end{figure}

            The Figure \ref{fig:binary_class_product_injection_SC1.png} presents the Performance Scores for the models—ResNet, MobileNet, EfficientNet, and VGG—across three approaches: Average Injection of Uncertainty (AIU), Direct Injection of Uncertainty (DIU) using Product Injection method, and a baseline. These scores elucidate the impact of uncertainty injection methodologies on model optimization. The baseline scores establish a reference point against which the efficacy of the AIU and DIU approaches can be evaluated. High-performance scores indicate superior performance. Notably, the results demonstrate that AIU consistently outperforms DIU for some models, particularly ResNet and EfficientNet. For instance, ResNet achieves a Performance Score of \textit{-0.40529} with AIU, compared to \textit{-0.40599} with DIU, indicating a marginal improvement in performance. Conversely, MobileNet exhibits a better Performance Score with DIU at \textit{-0.40616} compared to \textit{-0.40728} with AIU, suggesting that, in this case, the Direct Injection of Uncertainty is more effective. This trend is also reflected in VGG, where the AIU scores are less favorable than those obtained via DIU. Overall, both methods demonstrate superior performance compared to the baseline scores, highlighting the effectiveness of uncertainty injection techniques in optimizing model training across various architectures.

        \subsubsection{Deep Injection of Uncertainty}
            The Deep Injection of Uncertainty refines the model’s training by scaling the classification loss through the multiplication factor $w(K)$, now determined from Scorecard A. Performance results were obtained via the Deep Injection methodology, based on the origin of the Multiplication Factor—whether from DIU or AIU. These results are detailed in the corresponding summary \Cref{tab:Model Performance Scores for DIU and AIU- Inside Mul. Approach - 1SC}.
            \captionsetup{skip=10pt}
            \begin{table}[H] 
                \centering
                \caption{Model Performance Scores with DIU and AIU using Deep Injection of Uncertainty}
                \label{tab:Model Performance Scores for DIU and AIU- Inside Mul. Approach - 1SC}
                \resizebox{\columnwidth}{!}{%
                \begin{tabular}{|c|c|c|c|c|c|c|}
                                    \hline
                                    \textbf{Model} & \textbf{Epoch (DIU)} & \textbf{DIU} & \textbf{Epoch (AIU)} & \textbf{AIU} & \textbf{Epoch(Base.)} & \textbf{Baseline} \\
                                    \hline
                    ResNet & \textbf{12} & \textbf{-0.45286} & 14 & \textbf{-0.43328} & 13 & -0.47268 \\
                    \hline
                    MobileNet & 39 & \textbf{-0.45669} & 39 & \textbf{-0.43875} & 35 & -0.49351 \\
                    \hline
                    EfficientNet & \textbf{37} & \textbf{-0.34523} & \textbf{36} & \textbf{-0.33619} & 38 & -0.36971 \\
                    \hline
                    VGG & \textbf{15} & \textbf{-0.57100} & \textbf{20} & \textbf{-0.53110} & 24 & -0.58170 \\
                    \hline
                \end{tabular}%
                }
            \end{table}
            \begin{figure}[ht]
                \centering
                \includegraphics[width=0.45\textwidth,keepaspectratio]{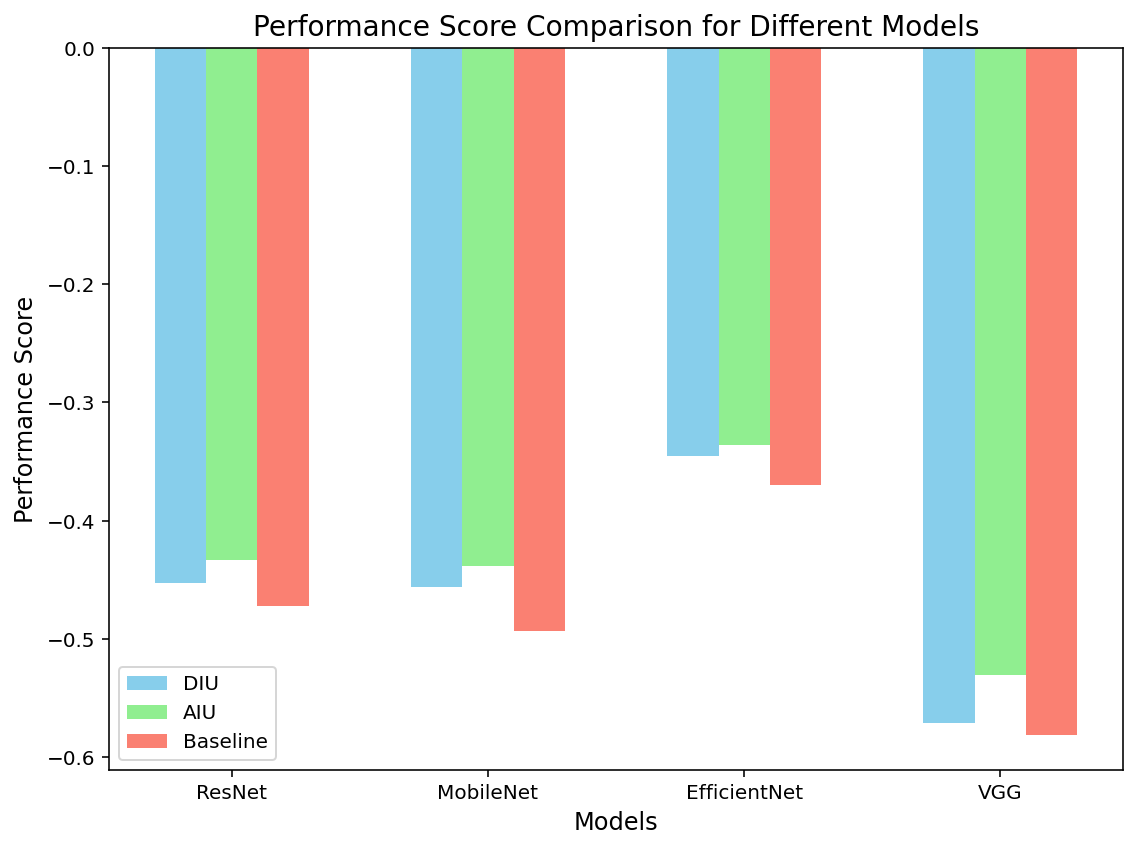}
                \caption{Deep Injection of Uncertainty - Evaluation of Model Performance Across DIU, AIU, and Baseline Approaches using Score Card A}
                \label{fig:binary_class_deep_injection_SC1.png}
            \end{figure}

            The figure \ref{fig:binary_class_deep_injection_SC1.png} illustrates the Performance Scores for ResNet, MobileNet, EfficientNet, and VGG across the Direct Injection of Uncertainty (DIU), Average Injection of Uncertainty (AIU), and baseline methodologies. This evaluation is conducted within a binary class framework, focusing on the single class "chair" to assess the Faster R-CNN model's effectiveness. The scores reveal that AIU consistently yields better values, indicating better performance compared to DIU for several models. For ResNet, the DIU score of \textit{-0.45286} is lower than the AIU score of \textit{-0.43328}, suggesting that AIU enhances training more effectively. MobileNet, EfficientNet, and VGG demonstrate a comparable trend, where the AIU yields better values compared to the DIU approach. Importantly, all DIU and AIU scores surpass their respective baseline scores. These findings establish the traditional training approach used to evaluate the Faster R-CNN model as a benchmark, highlighting the effectiveness of the Deep Injection of Uncertainty using the AIU approach in optimizing performance across various architectures.

    \subsection{Score Card A - Multi Class}            
        \subsubsection{Product Injection of Uncertainty}
            The Product Injection of Uncertainty extends its application to multi-class by adjusting the total loss across multiple classes using a multiplication factor $w(K)$, now derived from Scorecard A. As outlined in \Cref{subsec: Direct Injection of Uncertainty} and \Cref{subsec: Average Injection of Uncertainty}, the methodologies used to compute $M$ for the multi-class setting lead to diverse impacts on the model's Performance Score.

            Performance scores for the multi-class scenario, based on the source of the Multiplication Factor—namely, Direct Injection of Uncertainty (DIU) and Average Injection of Uncertainty (AIU)—are summarized in the table provided.
            \captionsetup{skip=10pt}
            \begin{table}[H] 
                \centering
                \caption{Model Performance Scores with DIU and AIU using Product Injection of Uncertainty - Four classes}
                \label{tab: Model Performance Scores for DIU and AIU- Multiplication Approach - 4 - 1SC}
                \resizebox{\columnwidth}{!}{%
                \begin{tabular}{|c|c|c|c|c|c|c|}
                                    \hline
                                    \textbf{Model} & \textbf{Epoch (DIU)} & \textbf{DIU} & \textbf{Epoch (AIU)} & \textbf{AIU} & \textbf{Epoch(Base.} & \textbf{Baseline} \\
                                    \hline
                    ResNet & 27 & -0.09903 & 36 & \textbf{-0.09416} & 16 & -0.09887 \\
                    \hline
                    MobileNet & \textbf{31} & \textbf{-0.09659} & \textbf{34} & -0.09963 & 36 & -0.09674 \\
                    \hline
                    EfficientNet & 38 & \textbf{0.02469} & 38 & \textbf{0.02601} & 37 & 0.02308 \\
                    \hline
                    VGG & \textbf{24} & \textbf{-0.25162} & \textbf{15} & -0.29528 & 28 & -0.26662 \\
                    \hline
                \end{tabular}%
                }
            \end{table}
            \begin{figure}[h]
                \centering
                \includegraphics[width=0.45\textwidth,keepaspectratio]{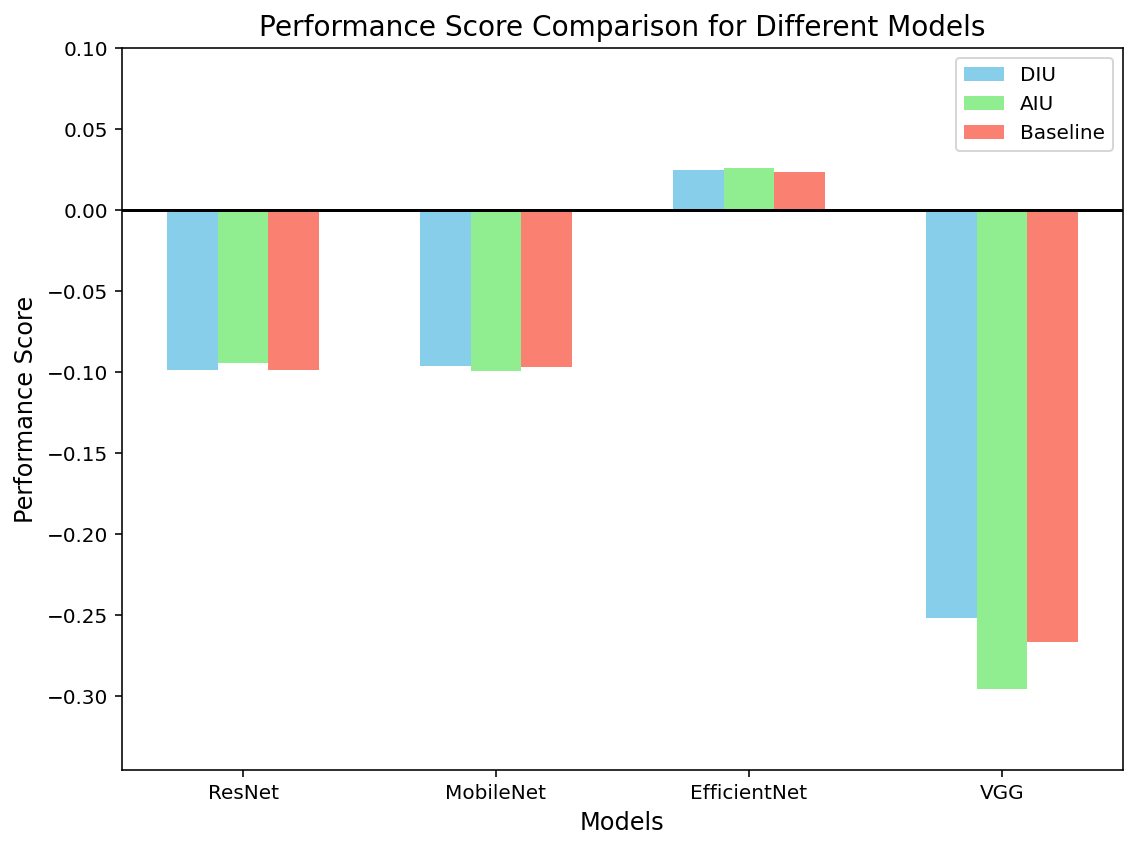}
                \caption{Product Injection of Uncertainty- Multiclass- Evaluation of Model Performance Across DIU, AIU, and Baseline Approaches using Score Card A}
                \label{fig:multi_class_product_injection_SC1.png}
            \end{figure}

            The figure \ref{fig:multi_class_product_injection_SC1.png} presents the Performance Scores for ResNet, MobileNet, EfficientNet, and VGG across the Product Injection of Uncertainty methodologies—Direct Injection of Uncertainty (DIU) and Average Injection of Uncertainty (AIU)—within a multi-class framework that includes the object classes "\textit{bus}", "\textit{car}", "\textit{motorbike}", and "\textit{train}". The performance scores indicate that AIU generally yields better results compared to DIU for most models. Specifically, MobileNet achieves an DIU score of \textit{-0.09659}, outperforming its AIU counterpart of \textit{-0.09963}. In the case of ResNet, the AIU score of \textit{-0.09416} is superior to the DIU score of \textit{-0.09903}, suggesting a more effective training enhancement through AIU. Although EfficientNet shows a slight advantage with AIU at \textit{0.02601} versus \textit{0.02469} for DIU, both scores remain close to each other. VGG, however, demonstrates a more pronounced difference, with a DIU score of \textit{-0.25162} significantly better than its AIU score of \textit{-0.29528}. Importantly, all DIU and AIU scores surpass their respective baseline scores, confirming the effectiveness of the Product Injection of Uncertainty methodology in optimizing model performance across multiple classes.

        \subsubsection{Deep Injection of Uncertainty}

            For the multi-class scenario, the Deep Injection of Uncertainty enhances the model's training by scaling the classification loss across all classes via the multiplication factor $w(K)$, which is now computed based on Scorecard A. Utilizing the Deep Injection approach, performance scores were obtained by evaluating the origin of the Multiplication Factor—either from DIU or AIU—as presented in the performance summary \Cref{tab:Model Performance Scores for DIU and AIU- Inside Mul. Approach - 4 - 1SC}.
            \captionsetup{skip=10pt}
            \begin{table}[H] 
                \centering
                \caption{Model Performance Scores with DIU and AIU using Deep Injection of Uncertainty - Four classes}
                 \label{tab:Model Performance Scores for DIU and AIU- Inside Mul. Approach - 4 - 1SC}
                \resizebox{\columnwidth}{!}{%
                \begin{tabular}{|c|c|c|c|c|c|c|}
                                    \hline
                                    \textbf{Model} & \textbf{Epoch (DIU)} & \textbf{DIU} & \textbf{Epoch (AIU)} & \textbf{AIU} & \textbf{Epoch(Base.} & \textbf{Baseline} \\
                                    \hline
                    ResNet & 26 & \textbf{-0.09036} & 25 & -0.10166 & 16 & -0.09887 \\
                    \hline
                    MobileNet & 39 & -0.11002 & 39 & -0.10804 & 36 & \textbf{-0.09674} \\
                    \hline
                    EfficientNet & 38 & \textbf{0.02497} & \textbf{36} & \textbf{0.02842} & 37 & 0.02308 \\
                    \hline
                    VGG & \textbf{17}& \textbf{-0.22602} & \textbf{12} & \textbf{-0.25242} & 28 & -0.26662 \\
                    \hline
                \end{tabular}%
                }
            \end{table}
            \begin{figure}[h]
                \centering
                \includegraphics[width=0.45\textwidth,keepaspectratio]{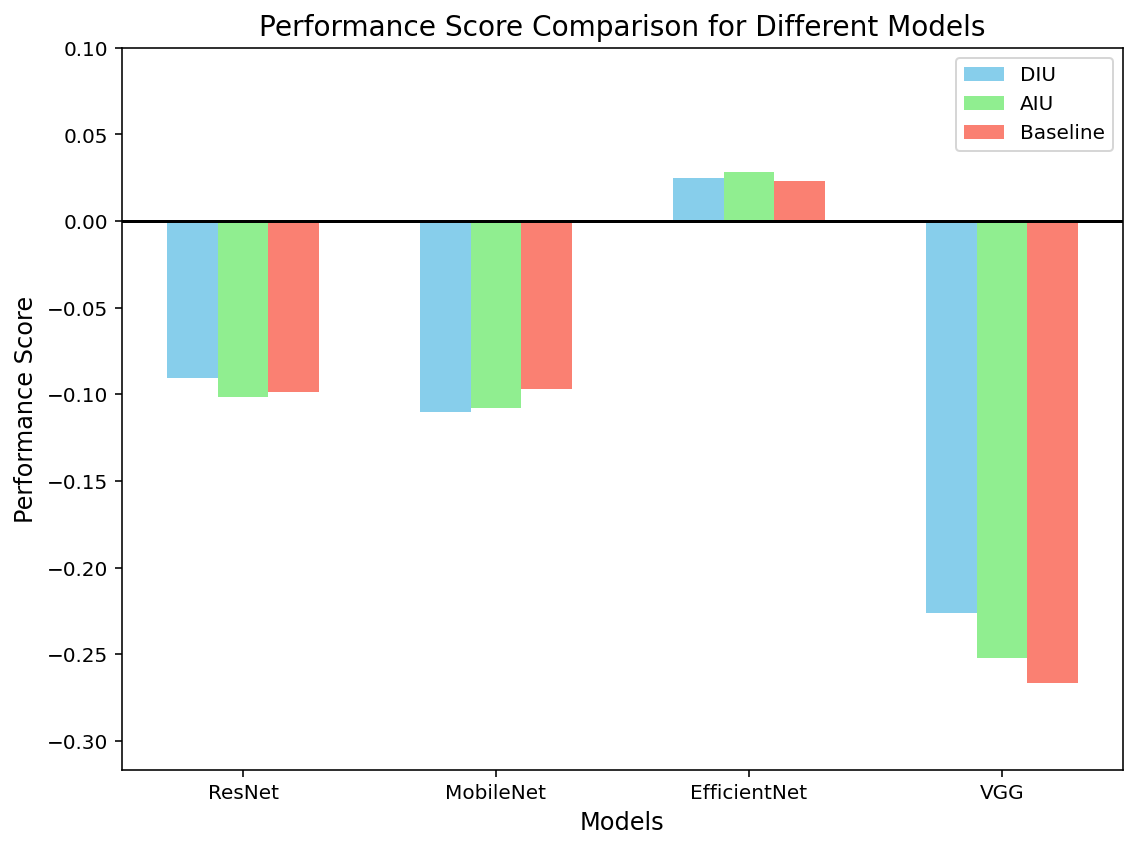}
                \caption{Deep Injection of Uncertainty- Multiclass- Evaluation of Model Performance Across DIU, AIU, and Baseline Approaches using Score Card A}
                \label{fig:multi_class_deep_injection_SC1.png}
            \end{figure}

            The figure \ref{fig:multi_class_deep_injection_SC1.png} illustrates the Performance Scores for ResNet, MobileNet, EfficientNet, and VGG using the Deep Injection of Uncertainty methodology in a multi-class framework that includes the object classes "\textit{bus}", "\textit{car}", "\textit{motorbike}", and "\textit{train}". The results indicate variability in scores across models, with DIU typically providing better performance than AIU. For example, ResNet exhibits a DIU score of \textit{-0.09036}, which surpasses the AIU score of \textit{-0.10166}, highlighting the effectiveness of DIU in enhancing training. In MobileNet, the DIU score of \textit{-0.11002} is slightly worse than the AIU score of \textit{-0.10804}. Meanwhile, EfficientNet shows comparable results, with a DIU score of \textit{0.02497} and an AIU score of \textit{0.02842}. VGG stands out with a DIU score of \textit{-0.22602}, outperforming its AIU score of \textit{-0.25242}. Notably, all DIU and AIU scores exceed their respective baseline scores, underscoring the effectiveness of the Deep Injection of Uncertainty methodology in optimizing model performance across multiple classes.

    \subsection{Score Card B - Binary Class}            
        \subsubsection{Product Injection of Uncertainty}
            The Product Injection of Uncertainty enhances the model's training process by scaling the total loss with multiplication factor $w(K)$ based on Scorecard B. As explained in \Cref{subsec: Direct Injection of Uncertainty} and \Cref{subsec: Average Injection of Uncertainty}, these methods provide distinct approaches to obtain $w(K)$, leading to different results in the model's Performance Score.
                
            The Performance Scores achieved using the Product Injection of Uncertainty, based on the origin of the Multiplication Factor value i.e., DIU and AIU are summarized in the following \Cref{tab: Model Performance Scores for DIU and AIU- Multiplication Approach}
            \captionsetup{skip=10pt}
            \begin{table}[H] 
                \centering
                \caption{Model Performance Scores with DIU and AIU Methods Using Product Injection of Uncertainty}
                \label{tab: Model Performance Scores for DIU and AIU- Multiplication Approach}
                \resizebox{\columnwidth}{!}{%
                \begin{tabular}{|c|c|c|c|c|c|c|}
                    \hline
                    \textbf{Model} & \textbf{Epoch (DIU)} & \textbf{DIU} & \textbf{Epoch (AIU)} & \textbf{AIU} & \textbf{Epoch(Base.} & \textbf{Baseline} \\
                    \hline
                    ResNet & \textbf{11} & \textbf{-0.32684} & \textbf{08} & \textbf{-0.34961} & 13 & -0.47268 \\
                    \hline
                    MobileNet & 38 & \textbf{-0.33678} & \textbf{28} & \textbf{-0.37383} & 35 & -0.49351 \\
                    \hline
                    EfficientNet & \textbf{38} & \textbf{-0.26775} & \textbf{36} & \textbf{-0.28221} & 38 & -0.36971 \\
                    \hline
                    VGG & \textbf{09} & \textbf{-0.39521} & \textbf{14} & \textbf{-0.40183} & 24 & -0.58170 \\
                    \hline
                \end{tabular}%
                }
            \end{table}
            \begin{figure}[h]
                \centering
                \includegraphics[width=0.45\textwidth,keepaspectratio]{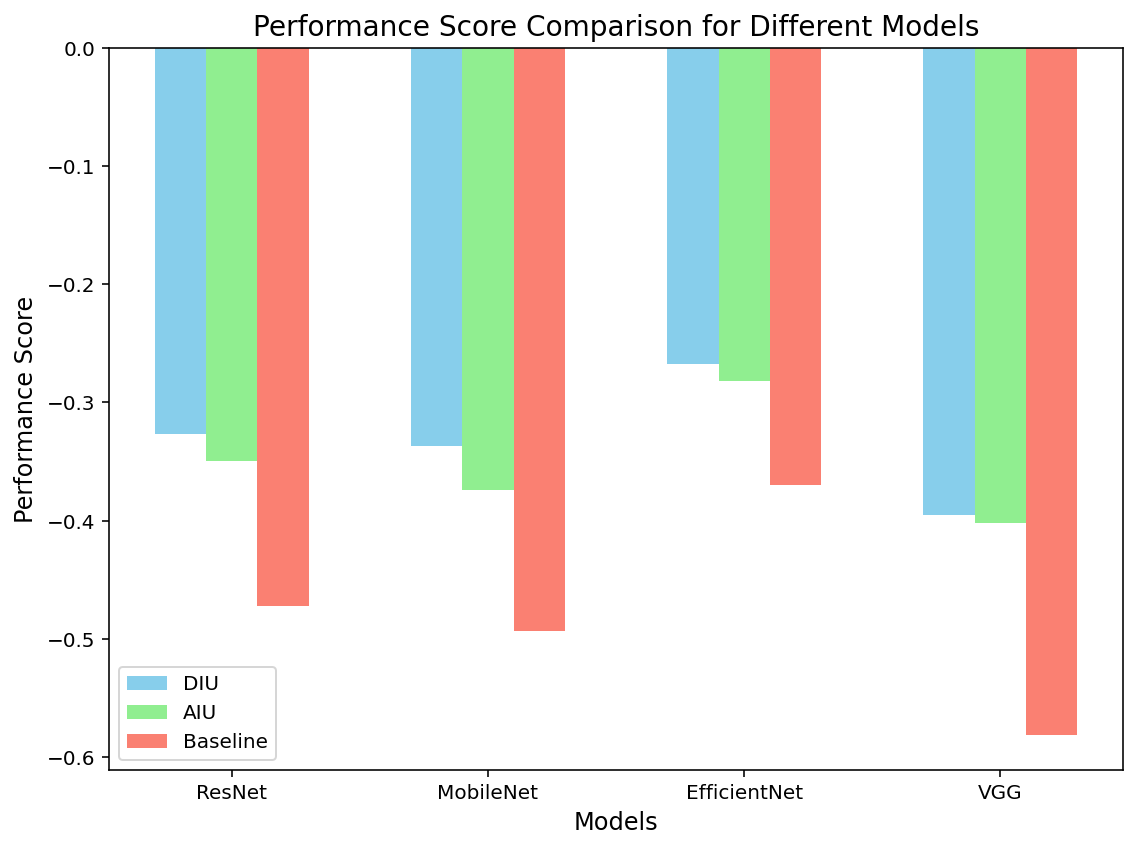}
                \caption{Product Injection of Uncertainty- Evaluation of Model Performance Across DIU, AIU, and Baseline Approaches using Score Card B}
                \label{fig:binary_class_product_injection_SC2.png}
            \end{figure}

            The Product Injection of Uncertainty using Scorecard B was evaluated across four models: ResNet, MobileNet, EfficientNet, and VGG, each designed to handle binary tasks. The performance scores achieved through this method indicate substantial enhancements over the baseline scores, as shown in the figure \ref{fig:binary_class_product_injection_SC2.png}.

            For instance, ResNet achieved a performance score of -0.32684 using Direct Injection of Uncertainty (DIU) and -0.34961 with Average Injection of Uncertainty (AIU), both reflecting improved performance compared to the baseline score of -0.47268. Similarly, MobileNet recorded scores of -0.33678 (DIU) and -0.37383 (AIU), outperforming its baseline of -0.49351. EfficientNet showed a performance score of -0.26775 with DIU and -0.28221 with AIU, both surpassing the baseline score of -0.36971. Finally, VGG's performance scores of -0.39521 (DIU) and -0.40183 (AIU) demonstrate a similar trend, as both methods improved its performance over the baseline score of -0.58170.

            Overall, the application of Product Injection of Uncertainty across these models confirms its effectiveness in enhancing their ability to generalize and accurately predict binary classes, thereby indicating the potential benefits of uncertainty integration in model training.

        \subsubsection{Deep Injection of Uncertainty}

            The Deep Injection of Uncertainty enhances the model's training process by scaling the classification loss with the multiplication factor $w(K)$, based on Scorecard B. The performance scores were achieved using the Inside Multiplication approach, based on the origin of the Multiplication factor value(i.e, DIU, and AIU), are summarized in the \Cref{tab:Model Performance Scores for DIU and AIU- Inside Mul. Approach}
            \captionsetup{skip=10pt}
            \begin{table}[H] 
                \centering
                \caption{Model Performance Scores with DIU and AIU using Deep Injection of Uncertainty}
                \label{tab:Model Performance Scores for DIU and AIU- Inside Mul. Approach}
                \resizebox{\columnwidth}{!}{%
                \begin{tabular}{|c|c|c|c|c|c|c|}
                                    \hline
                                    \textbf{Model} & \textbf{Epoch (DIU)} & \textbf{DIU} & \textbf{Epoch (AIU)} & \textbf{AIU} & \textbf{Epoch(Base.} & \textbf{Baseline} \\
                                    \hline
                    ResNet & 17 & \textbf{-0.41153} & 14 & \textbf{-0.43423} & 13 & -0.47268 \\
                    \hline
                    MobileNet & 37 & \textbf{-0.40426} & \textbf{35} & \textbf{-0.42654} & 35 & -0.49351 \\
                    \hline
                    EfficientNet & \textbf{36} & \textbf{-0.31584} & \textbf{35} & \textbf{-0.29313} & 38 & -0.36971 \\
                    \hline
                    VGG & \textbf{10} & \textbf{-0.51668} & \textbf{20} & \textbf{-0.52956} & 24 & -0.58170 \\
                    \hline
                \end{tabular}%
                }
            \end{table}
            
            \begin{figure}[h]
                \centering
                \includegraphics[width=0.45\textwidth,keepaspectratio]{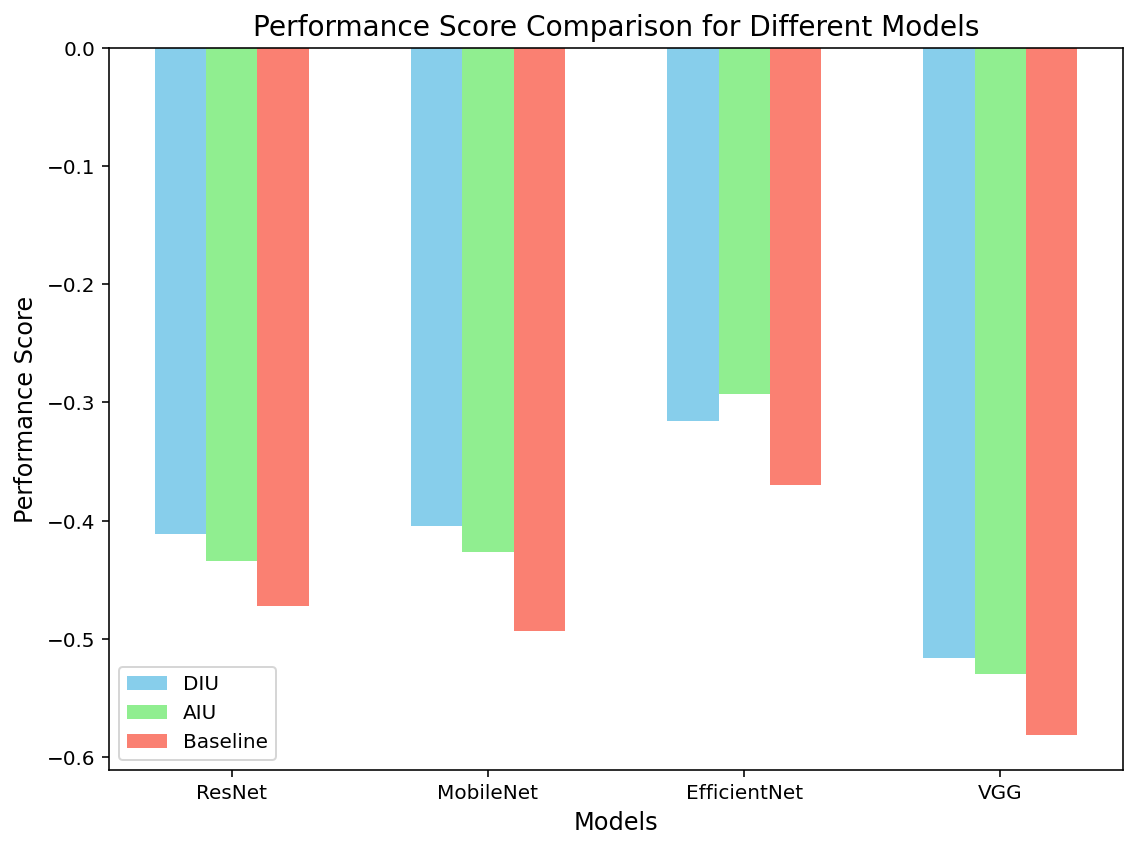}
                \caption{Deep Injection of Uncertainty- Evaluation of Model Performance Across DIU, AIU, and Baseline Approaches using Score Card B}
                \label{fig:binary_class_deep_injection_SC2.png}
            \end{figure}

            The Figure \ref{fig:binary_class_deep_injection_SC2.png} illustrates a comparison of performance across four models—ResNet, MobileNet, EfficientNet, and VGG. It reveals distinct advantages for each model when applying DIU and AIU methods. ResNet demonstrates a performance score of \textit{-0.41153} with DIU and \textit{-0.43423} with AIU. The higher DIU performance score indicates that ResNet benefits more compared to the AIU approach.
            
            MobileNet follows a similar trend, recording performance scores of \textit{-0.40426} with DIU and \textit{-0.42654} with AIU. Again, DIU shows superiority, suggesting that MobileNet's architecture effectively accommodates this method of uncertainty scaling. This consistent performance of DIU across both ResNet and MobileNet indicates that these architectures can leverage direct uncertainty integration to achieve better results.
            
            In contrast, EfficientNet's performance reflects a different dynamic, with scores of \textit{-0.31584} for DIU and \textit{-0.29313} for AIU. Here, AIU outperforms DIU, indicating that EfficientNet may derive more benefit from the average approach to uncertainty scaling. VGG presents a similar scenario to ResNet and MobileNet, achieving performance scores of \textit{-0.51668} with DIU and \textit{-0.52956} with AIU. Once again, DIU yields better results, highlighting that even a traditionally complex architecture like VGG can benefit from direct uncertainty integration.
            
            Overall, the comparative analysis reveals that while both DIU and AIU methods enhance model performance compared to baseline scores, the effectiveness of each method varies by model architecture. Specifically, DIU tends to perform better for ResNet, MobileNet, and VGG, while EfficientNet shows a preference for the AIU method.

    \subsection{Score Card B - Multi Class}            
        \subsubsection{Product Injection of Uncertainty}
            Building on the previous analysis for a single class, the Product Injection of Uncertainty was also applied to a broader scenario involving four classes. This adjustment was made to evaluate the scalability and consistency of the method across more complex datasets. The same DIU (Direct Injection of Uncertainty) and AIU (Average Injection of Uncertainty) methods were employed to derive the multiplication factor $w(K)$, with results reflecting the model's ability to generalize across multiple categories.
            \captionsetup{skip=10pt}
            \begin{table}[ht] 
                \centering
                \caption{Model Performance Scores with DIU and AIU using Product Injection of Uncertainty - Four classes}
                \label{tab: Model Performance Scores for DIU and AIU- Multiplication Approach - 4}
                \resizebox{\columnwidth}{!}{%
                \begin{tabular}{|c|c|c|c|c|c|c|}
                                    \hline
                                    \textbf{Model} & \textbf{Epoch (DIU)} & \textbf{DIU} & \textbf{Epoch (AIU)} & \textbf{AIU} & \textbf{Epoch(Base.} & \textbf{Baseline} \\
                                    \hline
                    ResNet & 22 & \textbf{-0.05346} & 28 & \textbf{-0.06670} & 16 & -0.09887 \\
                    \hline
                    MobileNet & 38 & \textbf{-0.04508} & 39 & \textbf{-0.04289} & 36 & -0.09674 \\
                    \hline
                    EfficientNet & 38 & \textbf{0.06719} & 39 & \textbf{0.07256} & 37 & 0.02308 \\
                    \hline
                    VGG & \textbf{13} & \textbf{-0.17987} & \textbf{20} & \textbf{-0.19373} & 28 & -0.26662\\
                    \hline
                \end{tabular}%
                }
            \end{table}
            
            \begin{figure}[ht]
                \centering
                \includegraphics[width=0.45\textwidth,keepaspectratio]{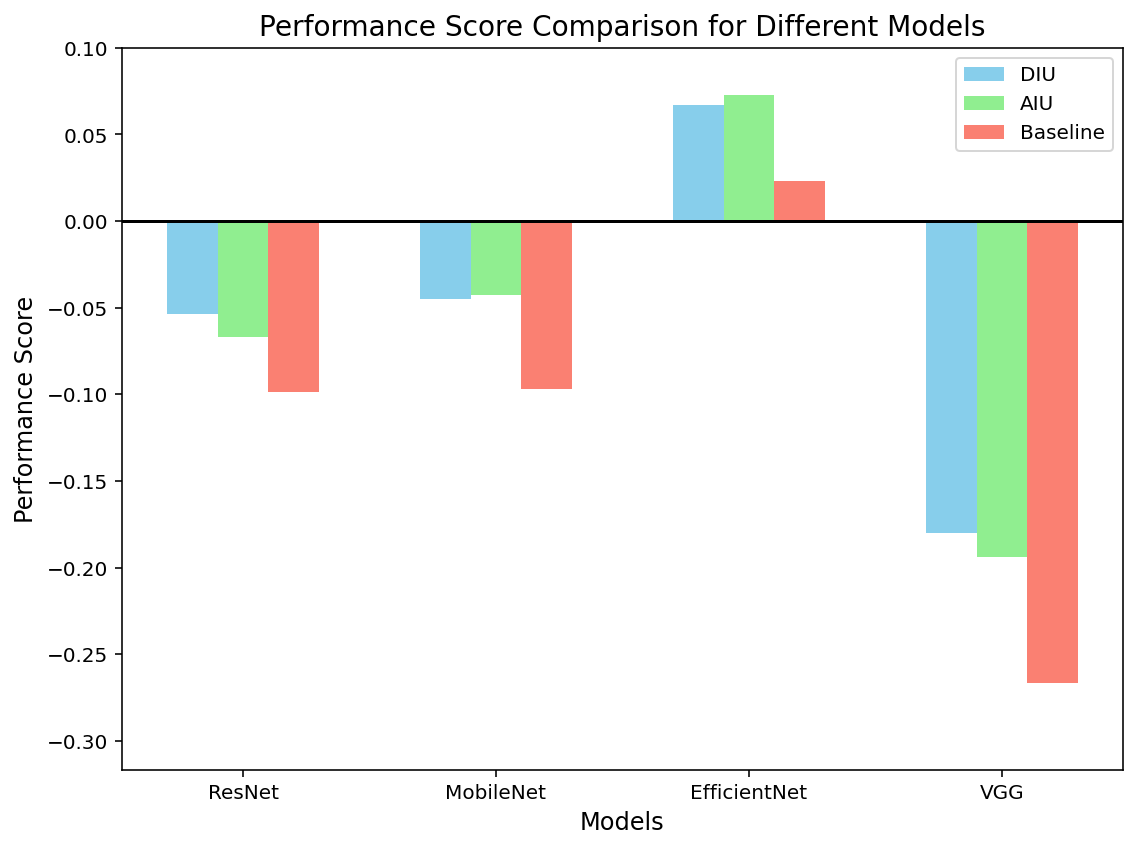}
                \caption{Product Injection of Uncertainty-Multi Class- Evaluation of Model Performance Across DIU, AIU, and Baseline Approaches using Score Card B}
                \label{fig: Multi_class_product_injection_SC2.png}
            \end{figure}

            The results in figure \ref{fig: Multi_class_product_injection_SC2.png} showcase the performance of four different models—ResNet, MobileNet, EfficientNet, and VGG—when trained using the Product Injection of Uncertainty approach across four object classes: "bus," "car," "motorbike," and "train." The multiplication factor $w(K)$, derived from Scorecard B, was applied to scale the total loss during the model's training process.

            For ResNet, the Direct Injection of Uncertainty (DIU) results in a performance score of \textit{-0.05346}, while the Average Injection of Uncertainty (AIU) score is slightly lower at \textit{-0.06670}, both demonstrating significant improvement over the baseline score of \textit{-0.09887}. MobileNet, however, shows a reversed pattern where AIU marginally outperforms DIU, with scores of \textit{-0.04289} and \textit{-0.04508}, respectively. The EfficientNet model, which consistently tends to yield positive values, performs best with AIU, scoring \textit{0.07256}, while DIU results in a marginally lower score of \textit{0.06719}. Lastly, VGG exhibits the largest improvement with DIU \textit{-0.17987} compared to AIU \textit{-0.19373}, both outperforming the baseline of \textit{-0.26662}.

            Overall, Product Injection of Uncertainty significantly enhances performance across all models, as both DIU and AIU consistently outperform their respective baselines. While the difference between DIU and AIU varies depending on the model architecture.

        \subsubsection{Deep Injection of Uncertainty}            
            
            The Deep Injection of Uncertainty for four-class training scales the classification loss using a multiplication factor, M, tailored to improve model performance across multiple categories. This approach adjusts the loss during training based on two distinct methods for deriving the multiplication factor: Direct Injection of Uncertainty (DIU) and Average Injection of Uncertainty (AIU). The performance scores achieved through these methods reflect the models' ability to generalize and maintain accuracy across multiple categories.

            \captionsetup{skip=10pt}
            \begin{table}[ht] 
                \centering
                \caption{Model Performance Scores with DIU and AIU using Deep Injection of Uncertainty - Four classes}
                 \label{tab:Model Performance Scores for DIU and AIU- Inside Mul. Approach - 4}
                \resizebox{\columnwidth}{!}{%
                \begin{tabular}{|c|c|c|c|c|c|c|}
                                    \hline
                                    \textbf{Model} & \textbf{Epoch (DIU)} & \textbf{DIU} & \textbf{Epoch (AIU)} & \textbf{AIU} & \textbf{Epoch(Base.} & \textbf{Baseline} \\
                                    \hline
                     ResNet & 31 & \textbf{-0.07168} & 19 & \textbf{-0.08424} & 16 & -0.09887 \\
                    \hline
                    MobileNet & 38 & \textbf{-0.06784} & \textbf{35} & \textbf{-0.08124} & 36 & -0.09674 \\
                    \hline
                    EfficientNet & \textbf{35} & \textbf{0.06308} & 38 & \textbf{0.04832} & 37 & 0.02308 \\
                    \hline
                    VGG & \textbf{24} & \textbf{-0.26209} & \textbf{11} & \textbf{-0.20404} & 28 & -0.26662\\
                    \hline
                \end{tabular}%
                }
            \end{table}
            
            \begin{figure}[ht]
                \centering
                \includegraphics[width=0.45\textwidth,keepaspectratio]{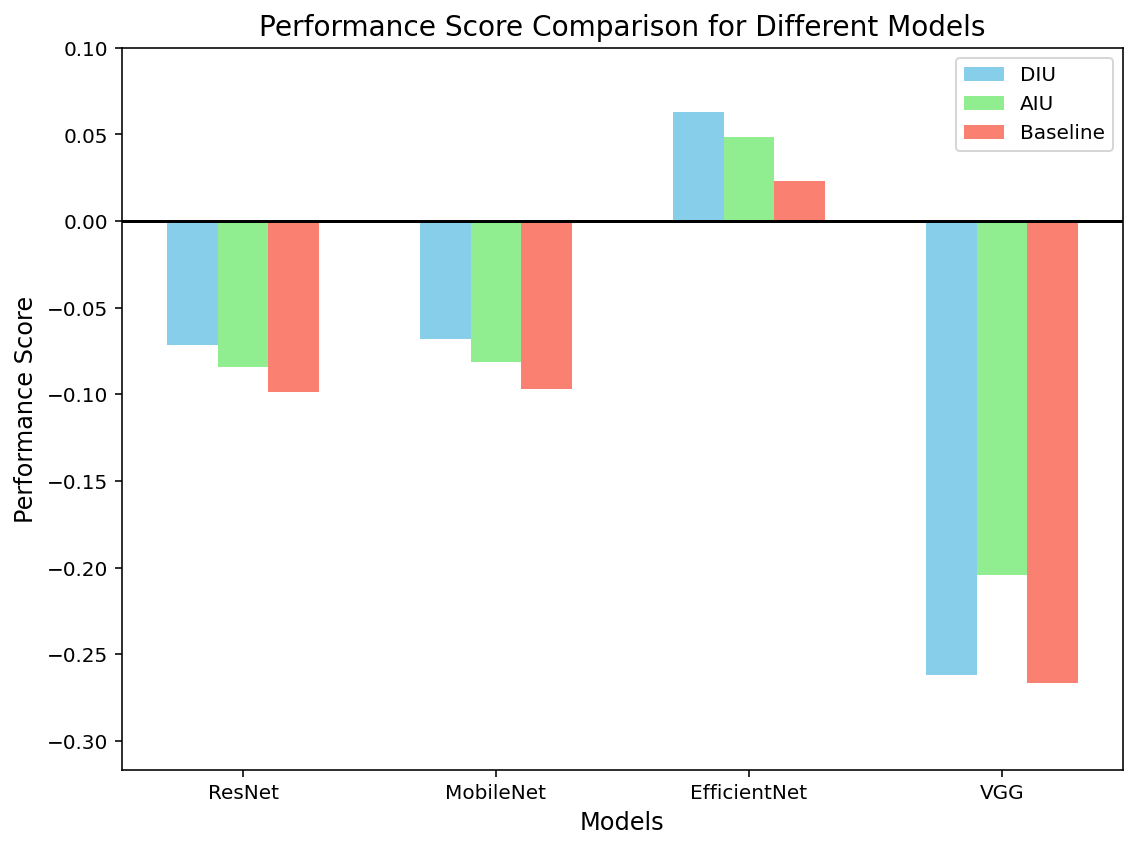}
                \caption{Deep Injection of Uncertainty-Multi Class- Evaluation of Model Performance Across DIU, AIU, and Baseline Approaches using Score Card B}
                \label{fig: Multi_class_deep_injection_SC2.png}
            \end{figure}
            
            The results in \Cref{fig: Multi_class_deep_injection_SC2.png} evaluate the same set of models—ResNet, MobileNet, EfficientNet, and VGG—trained using the Deep Injection of Uncertainty approach across the same four object classes. Unlike Product Injection, Deep Injection of Uncertainty more deeply integrates uncertainty scaling into the model’s training process by adjusting the classification loss using multiplication factor $w(K)$ from Scorecard B.
            
            In ResNet, Deep Injection of Uncertainty leads to a performance score of \textit{-0.07168} with DIU and \textit{-0.08424} with AIU, showing that both methods improve significantly over the baseline score of \textit{-0.09887}, with DIU yielding the better outcome. MobileNet shows similar behavior, with DIU (\textit{-0.06784}) once again outperforming AIU (\textit{-0.08124}), although both approaches still improve on the baseline score of \textit{-0.09674}. EfficientNet also behaves similarly in this case, with DIU yielding a better performance score (\textit{0.06308}) compared to AIU (\textit{0.04832}), though both still surpass the baseline score of \textit{0.02308}. In the case of VGG, both DIU (\textit{-0.26209}) and AIU (\textit{-0.20404}) outperform the baseline (\textit{-0.26662}), with AIU providing better results.

    \subsection{Results and Discussion}

        This section evaluates object detection models using different uncertainty injection methods \textit{Direct Injection of Uncertainty (DIU)} and \textit{Average Injection of Uncertainty (AIU)} across single-class and multi-class datasets. Two approaches for incorporating uncertainty into the loss function are compared: \textit{Product Injection of Uncertainty} and \textit{Deep Injection of Uncertainty}, utilizing \textit{Score Card A} and \textit{Score Card B}.
        
        Integrating uncertainty into the training process consistently enhances model performance over the baseline. In single-class detection, AIU generally outperforms DIU for models like ResNet and EfficientNet, suggesting that averaging uncertainty provides smoother loss adjustments leading to better performance. Conversely, models like MobileNet and VGG benefit more from DIU, indicating that direct scaling of the loss function is more effective for certain architectures.
        
        In multi-class detection involving categories like "bus," "car," "motorbike," and "train," uncertainty integration continues to improve performance. With \textbf{Product Injection of Uncertainty} and \textbf{Score Card A}, AIU often yields better results for models like ResNet and EfficientNet, while VGG shows improved performance with DIU. This variation suggests that the optimal uncertainty injection method may depend on the model architecture.
        
        Using \textbf{Score Card B}, DIU often provides better performance across both single-class and multi-class experiments. This indicates that the choice of scorecard and uncertainty injection method should be tailored to the specific model and dataset for optimal results.
        
        Overall, \textbf{Product Injection of Uncertainty combined with Direct Injection of Uncertainty (DIU)} generally offers the best performance across various models and datasets. This combination enhances stability through smoother loss adjustments and effectively integrates uncertainty into the training process. Models like ResNet and EfficientNet benefit the most from this approach due to their balanced complexity and depth.
        
        
        In conclusion, integrating uncertainty management, particularly through Product Injection of Uncertainty with DIU provides a robust strategy for improving object detection models' performance and training efficiency. This methodology holds significant promise for applications requiring high accuracy and reliability, such as autonomous vehicles and surveillance systems.

\section{Full Dataset Experiment}
Based on the findings from previous experiments and results, the approach selected for further analysis is Direct Injection of Uncertainty (DIU) combined with Product Injection of Uncertainty, and Score card B was used for the Multiplication factor.

This combination was chosen due to its demonstrated effectiveness in enhancing model training efficiency and accuracy. The next evaluation phase involves applying this method to the entire Pascal VOC dataset to determine whether the advantages observed in earlier tests on smaller subsets continue to hold when scaling to a comprehensive, diverse dataset.

The analysis focuses on using Direct Injection of Uncertainty within the Product Injection framework, where uncertainty is directly injected into the training process. This approach dynamically scales the overall loss function, considering uncertainty values derived from the Dempster-Shafer Theory. By leveraging uncertainty-weighted feedback, the training process becomes more adaptive, prioritizing ambiguous predictions and refining model parameters more effectively.

The results from the Direct Injection of Uncertainty combined with Product Injection will be directly compared against those from the traditional training methods. This comparison aims to validate whether the novel approach continues to deliver superior performance, as seen in earlier experiments, particularly in terms of both precision and training efficiency.

\subsection{Performance Results}

The results of applying the Direct Injection of Uncertainty (DIU) combined with Product Injection of Uncertainty on the full Pascal VOC dataset, which contains 20 object classes, demonstrate significant improvements in model performance compared to traditional training methods as shown in \Cref{tab: Model Performance Scores for DIU Multiplication Approach - 20}. This section systematically evaluates the impact of integrating uncertainty-based training on various FasterRCNN models with different backbones, namely ResNet, MobileNet, EfficientNet, and VGG.

\captionsetup{skip=10pt}
            \begin{table}[ht] 
                \centering
                \caption{Model Performance Scores with DIU using Product Injection of Uncertainty - 20 classes}
                \label{tab: Model Performance Scores for DIU Multiplication Approach - 20}
                \resizebox{\columnwidth}{!}{%
                \begin{tabular}{|c|c|c|c|c|c|c|}
                                    \hline
                                    \textbf{Model} & \textbf{Epoch (DIU)} & \textbf{DIU}  & \textbf{Epoch(Base.} & \textbf{Baseline} \\
                                    \hline
                    ResNet & \textbf{23} & \textbf{-0.30428} & 30 & -0.35063 \\
                    \hline
                    MobileNet & \textbf{37} & \textbf{-0.35934} & 38 & -0.43241 \\
                    \hline
                    EfficientNet & \textbf{37} & \textbf{-0.22646} & 38 & -0.28324 \\
                    \hline
                    VGG & \textbf{13} & \textbf{-0.47960} & 17 & -0.50840\\
                    \hline
                \end{tabular}%
                }
            \end{table}

            \begin{figure}[H]
                \centering
                \includegraphics[width=0.45\textwidth,keepaspectratio]{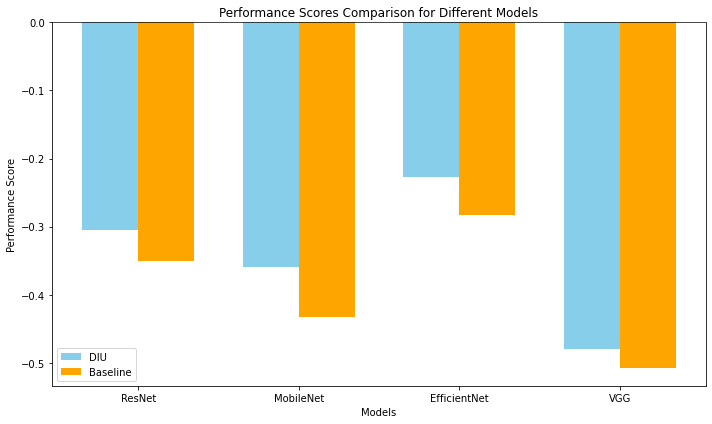}
                \caption{Product Injection of Uncertainty - Evaluation of Model Performance Across DIU and Baseline Approaches using Score Card B}
                \label{fig:20classes}
            \end{figure}
The results from \Cref{tab: Model Performance Scores for DIU Multiplication Approach - 20} shows that incorporating uncertainty into the training process using DIU leads to substantial gains in performance metrics across all tested models which was also seen in the smaller dataset experiments. For instance, ResNet achieved a performance score of \textit{-0.30075}, notably better than its baseline performance score of \textit{-0.35125}. Similarly, MobileNet improved its performance from a baseline of \textit{-0.43199} to \textit{-0.36641} with the DIU approach. The EfficientNet model showed an improvement in its performance score from \textit{-0.29175} to \textit{-0.22642}, indicating that the uncertainty-based adjustments enhanced its prediction accuracy. Even the VGG model, which traditionally faces challenges in optimizing object detection tasks due to its complexity, achieved a score of \textit{-0.46632}, compared to its baseline of \textit{-0.50338}.

The scalability of this uncertainty-driven training method was validated by applying it to the full Pascal VOC dataset, which is diverse and challenging due to its inclusion of 20 distinct object classes. The fact that the DIU approach continued to outperform traditional training methods even on this comprehensive dataset indicates its robustness and scalability. This confirms that leveraging uncertainty in training not only improves performance in smaller subsets but also remains effective in more complex, real-world datasets, making it a promising strategy for large-scale object detection tasks.

Overall, the experiment highlights the benefits of integrating Direct Injection of Uncertainty with Product Injection, offering a substantial enhancement in both model accuracy and training efficiency. This approach effectively demonstrates how uncertainty-based methodologies can optimize object detection frameworks, paving the way for their application in more advanced and diverse AI systems.

\subsection{Mean Average Precision (mAP) Report}
The Mean Average Precision (mAP) scores for the Direct Injection of Uncertainty (DIU) method compared to the baseline models across various architectures are presented in Table~\ref{tab:merged_map}. The mAP metric provides a comprehensive evaluation of the model's precision across all object classes, serving as a crucial indicator of detection performance.
\begin{table}[H]
    \centering
    \resizebox{\columnwidth}{!}{
    \begin{tabular}{|l|cc|cc|cc|cc|}
        \hline
        \textbf{Class} & \multicolumn{2}{c|}{\textbf{ResNet}} & \multicolumn{2}{c|}{\textbf{MobileNet}} & \multicolumn{2}{c|}{\textbf{EfficientNet}} & \multicolumn{2}{c|}{\textbf{VGG}} \\ \hline
        & \textbf{Baseline} & \textbf{DIU} & \textbf{Baseline} & \textbf{DIU} & \textbf{Baseline} & \textbf{DIU} & \textbf{Baseline} & \textbf{DIU} \\ \hline
        Person         & 0.35 & 0.34 & 0.33 & 0.33 & 0.38 & 0.38 & 0.32 & 0.33 \\ 
        Bird           & 0.26 & 0.25 & 0.24 & 0.23 & 0.33 & 0.31 & 0.22 & 0.20 \\ 
        Cat            & 0.39 & 0.40 & 0.36 & 0.36 & 0.43 & 0.42 & 0.33 & 0.32 \\ 
        Cow            & 0.29 & 0.28 & 0.25 & 0.28 & 0.31 & 0.32 & 0.22 & 0.21 \\ 
        Dog            & 0.35 & 0.36 & 0.32 & 0.32 & 0.40 & 0.39 & 0.29 & 0.29 \\ 
        Horse          & 0.18 & 0.17 & 0.17 & 0.17 & 0.27 & 0.28 & 0.14 & 0.13 \\ 
        Sheep          & 0.31 & 0.29 & 0.24 & 0.27 & 0.30 & 0.29 & 0.26 & 0.26 \\ 
        Aeroplane      & 0.35 & 0.35 & 0.34 & 0.34 & 0.41 & 0.39 & 0.31 & 0.31 \\ 
        Bicycle        & 0.28 & 0.28 & 0.25 & 0.24 & 0.28 & 0.23 & 0.21 & 0.23 \\ 
        Boat           & 0.11 & 0.12 & 0.13 & 0.12 & 0.17 & 0.15 & 0.11 & 0.09 \\ 
        Bus            & 0.41 & 0.40 & 0.43 & 0.41 & 0.44 & 0.44 & 0.38 & 0.37 \\ 
        Car            & 0.26 & 0.26 & 0.23 & 0.24 & 0.27 & 0.26 & 0.22 & 0.24 \\ 
        Motorbike      & 0.30 & 0.29 & 0.32 & 0.32 & 0.36 & 0.32 & 0.26 & 0.25 \\ 
        Train          & 0.29 & 0.28 & 0.31 & 0.29 & 0.38 & 0.37 & 0.27 & 0.26 \\ 
        Bottle         & 0.14 & 0.13 & 0.12 & 0.11 & 0.13 & 0.14 & 0.10 & 0.10 \\ 
        Chair          & 0.09 & 0.09 & 0.07 & 0.07 & 0.10 & 0.09 & 0.07 & 0.07 \\ 
        Dining Table   & 0.17 & 0.15 & 0.18 & 0.19 & 0.19 & 0.20 & 0.17 & 0.15 \\ 
        Potted Plant   & 0.09 & 0.08 & 0.08 & 0.08 & 0.09 & 0.10 & 0.05 & 0.05 \\ 
        Sofa           & 0.21 & 0.24 & 0.22 & 0.23 & 0.25 & 0.26 & 0.17 & 0.17 \\ 
        TV Monitor     & 0.27 & 0.27 & 0.27 & 0.24 & 0.29 & 0.28 & 0.24 & 0.23 \\ \hline\hline
        \textbf{Global} & \textbf{0.25} & \textbf{0.25} & \textbf{0.25} & \textbf{0.25} & \textbf{0.29} & \textbf{0.28} & \textbf{0.22} & \textbf{0.22} \\ \hline
    \end{tabular}
    }
    \caption{Mean Average Precision (mAP) of DIU with respect to Baseline across different models}
    \label{tab:merged_map}
\end{table}
The DIU method demonstrates competitive performance, often matching or exceeding the baseline, such as in the 'Cat' class (ResNet: 0.40 vs. 0.39) and 'Sofa' class (ResNet: 0.24 vs. 0.21). However, the baseline outperforms DIU in some cases, like the 'Bird' class (EfficientNet: 0.33 vs. 0.31). Global mAP scores remain comparable, with both DIU and baseline achieving 0.25 for ResNet and MobileNet, while EfficientNet and VGG show minor differences.

These results highlight DIU's viability as a robust alternative for object detection, with consistent performance across architectures and classes.

\subsection{Confusion Matrices and F1 Scores}

Confusion matrices and F1 scores are crucial for evaluating the performance of classification models in multi-class problems, such as those encountered in the Pascal VOC dataset. The Pascal VOC dataset includes 20 object classes and uses a multi-class framework where each class is evaluated separately and overall metrics are aggregated.

A multi-class confusion matrix is a square matrix of size \(N \times N\), where \(N\) is the number of classes. Each row represents the actual class, while each column represents the predicted class. For Pascal VOC, the confusion matrix is a \(20 \times 20\) matrix, and each entry \((i, j)\) indicates the number of samples belonging to class \(i\) that were predicted as class \(j\).

For each class \(i\), we can calculate the precision, recall, and F1 score:

\begin{equation}    
    \text{Precision}_i = \frac{\text{TP}_i}{\text{TP}_i + \sum_{j \neq i} \text{FP}_{i,j}},
\end{equation}
\begin{equation}  
    \text{Recall}_i = \frac{\text{TP}_i}{\text{TP}_i + \sum_{j \neq i} \text{FN}_{j,i}}
\end{equation}

Here:
\begin{itemize}
    \item \(\text{TP}_i\): True Positives for class \(i\) (correctly predicted as class \(i\)).
    \item \(\text{FP}_{i,j}\): False Positives for class \(i\) (instances of other classes predicted as class \(i\)).
    \item \(\text{FN}_{j,i}\): False Negatives for class \(i\) (instances of class \(i\) predicted as other classes).
\end{itemize}

The F1 score for class \(i\) is then computed as:

\begin{equation}
    \text{F1}_i = 2 \cdot \frac{\text{Precision}_i \cdot \text{Recall}_i}{\text{Precision}_i + \text{Recall}_i}
\end{equation}

To evaluate the overall performance across all classes, micro-averaging is commonly used, especially in datasets like Pascal VOC where class imbalances may occur. The \textbf{Micro-Averaged F1 Score} aggregates the contributions of all classes by summing the true positives, false positives, and false negatives across all classes before computing precision and recall. The formula is given by:

\begin{equation}
    \text{Precision}_{\text{micro}} = \frac{\sum_{i=1}^N \text{TP}_i}{\sum_{i=1}^N (\text{TP}_i + \text{FP}_i)},
    \label{eq:Precision_micro}
\end{equation}
\begin{equation}
    \text{Recall}_{\text{micro}} = \frac{\sum_{i=1}^N \text{TP}_i}{\sum_{i=1}^N (\text{TP}_i + \text{FN}_i)},
    \label{eq:Recall_micro}
\end{equation}

The micro-averaged F1 score is then calculated as the harmonic mean of the micro-averaged precision and recall:

\begin{equation}
    \text{F1}_{\text{micro}} = 2 \cdot \frac{\text{Precision}_{\text{micro}} \cdot \text{Recall}_{\text{micro}}}{\text{Precision}_{\text{micro}} + \text{Recall}_{\text{micro}}}
    \label{eq:F1_micro}
\end{equation}

Micro-averaging is particularly useful for evaluating imbalanced datasets, as it gives equal importance to each prediction regardless of class size.

\subsubsection{ResNet:}

\begin{figure}[H]
    \centering
    \begin{subfigure}[b]{0.45\textwidth}
        \centering
        \includegraphics[width=\textwidth]{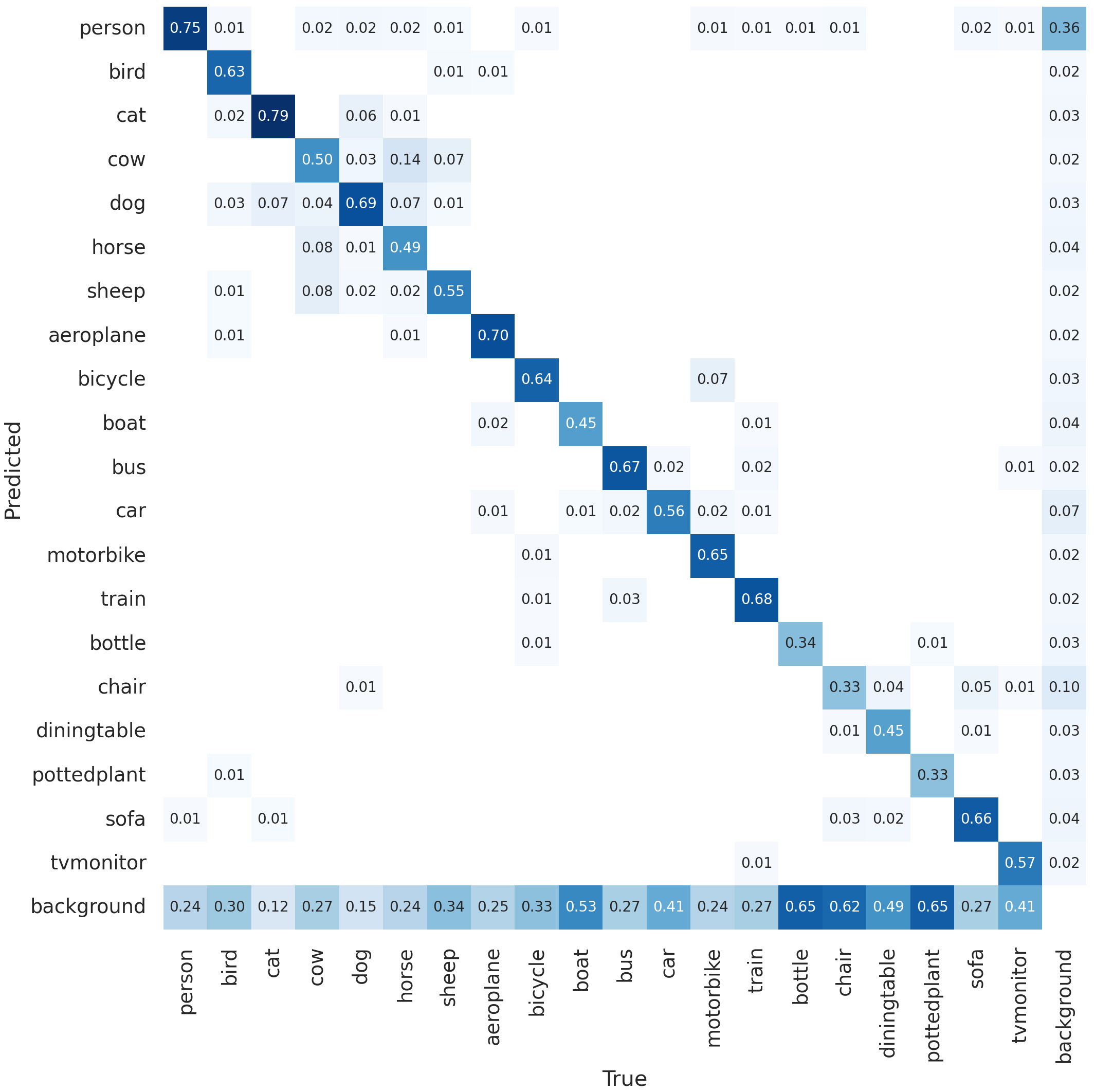}
        \caption{Confusion Matrix for Baseline at Epoch 30}
        \label{fig:resnet_cm_base}
    \end{subfigure}
    \\[1cm]
    \begin{subfigure}[b]{0.45\textwidth}
        \centering
        \includegraphics[width=\textwidth]{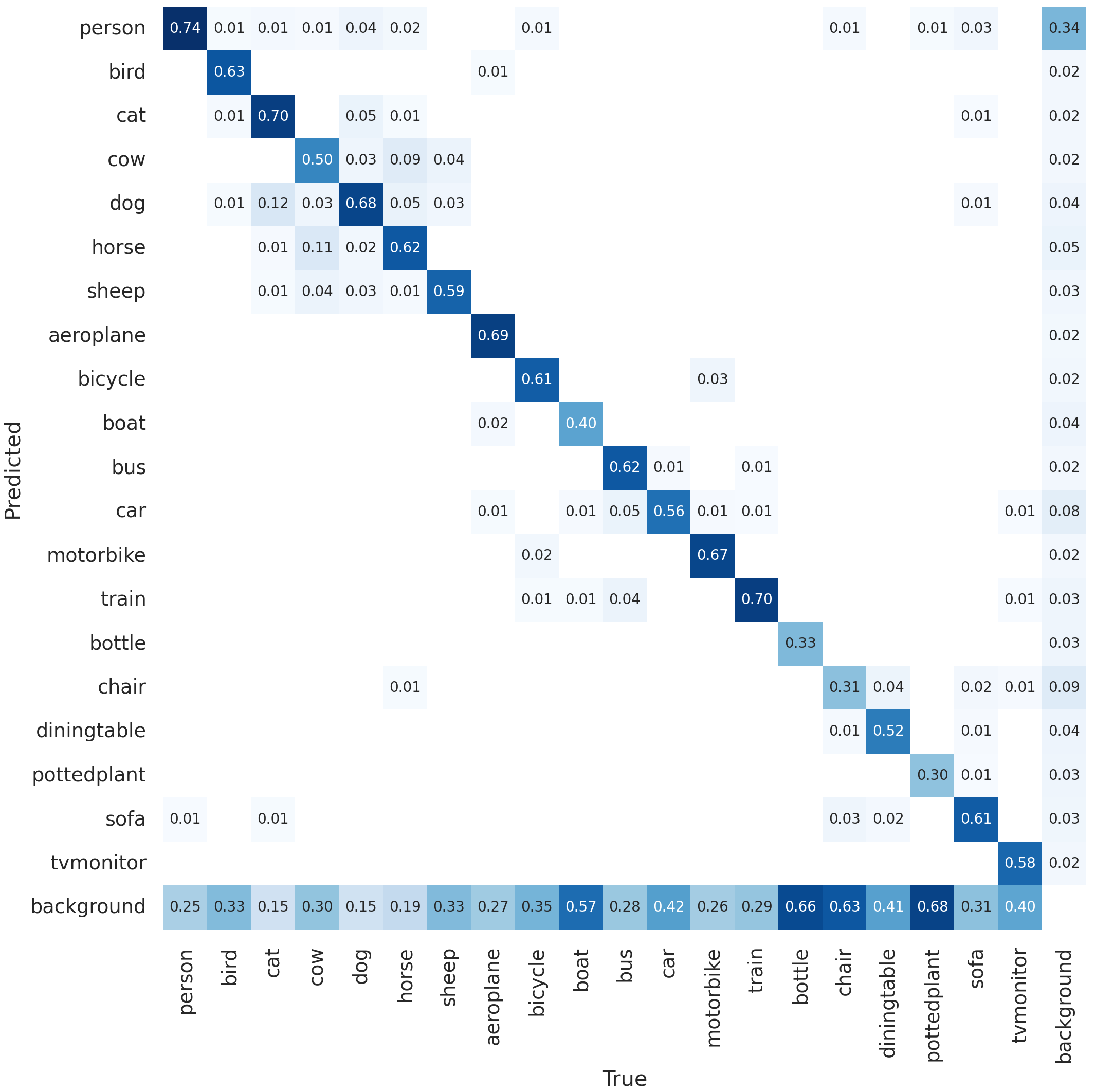}
        \caption{Confusion Matrix for DIU at Epoch 23}
        \label{fig:resnet_cm_diu}
    \end{subfigure}
    \caption{Confusion Matrices for ResNet Baseline vs DIU}
    \label{fig:resnet_cm}
\end{figure}

The classification performance of the ResNet model is analyzed using confusion matrices and evaluation metrics. Figure~\ref{fig:resnet_cm_base} shows the confusion matrix for the baseline at Epoch 30, while Figure~\ref{fig:resnet_cm_diu} shows the confusion matrix for the DIU method at Epoch 23. These confusion matrices highlight the performance differences in detecting various object classes.

Table~\ref{tab:resnet_metrics} presents the class-based metrics, including precision, recall, and F1 scores, for both the baseline and DIU methods. These metrics provide insights into the model's performance for individual classes and highlight the improvements achieved by the DIU method in specific cases.

\begin{table}[H]
    \centering
    \resizebox{\columnwidth}{!}{
    \begin{tabular}{|l|c|c|c|c|c|c|}
        \hline
        \textbf{Class} & \multicolumn{3}{|c|}{\textbf{Baseline}} & \multicolumn{3}{|c|}{\textbf{DIU}} \\
        \cline{2-7}
         & Precision & Recall & F1 Score & Precision & Recall & F1 Score \\
        \hline
        Person        & 0.52 & 0.76 & 0.62 & 0.51 & 0.76 & 0.61 \\
        Bird          & 0.53 & 0.64 & 0.58 & 0.52 & 0.65 & 0.58 \\
        Cat           & 0.55 & 0.77 & 0.64 & 0.60 & 0.74 & 0.66 \\
        Cow           & 0.34 & 0.53 & 0.41 & 0.33 & 0.46 & 0.38 \\
        Dog           & 0.42 & 0.71 & 0.53 & 0.50 & 0.59 & 0.54 \\
        Horse         & 0.21 & 0.55 & 0.30 & 0.22 & 0.60 & 0.33 \\
        Sheep         & 0.48 & 0.52 & 0.50 & 0.45 & 0.58 & 0.51 \\
        Aeroplane     & 0.56 & 0.68 & 0.61 & 0.51 & 0.66 & 0.57 \\
        Bicycle       & 0.34 & 0.66 & 0.45 & 0.34 & 0.62 & 0.44 \\
        Boat          & 0.28 & 0.45 & 0.35 & 0.32 & 0.44 & 0.37 \\
        Bus           & 0.40 & 0.67 & 0.50 & 0.42 & 0.65 & 0.51 \\
        Car           & 0.35 & 0.67 & 0.46 & 0.42 & 0.60 & 0.50 \\
        Motorbike     & 0.36 & 0.67 & 0.47 & 0.41 & 0.61 & 0.49 \\
        Train         & 0.27 & 0.31 & 0.29 & 0.23 & 0.34 & 0.27 \\
        Bottle        & 0.26 & 0.34 & 0.30 & 0.30 & 0.42 & 0.35 \\
        Chair         & 0.31 & 0.32 & 0.31 & 0.30 & 0.36 & 0.33 \\
        Dining Table  & 0.32 & 0.36 & 0.34 & 0.30 & 0.40 & 0.34 \\
        Potted Plant  & 0.31 & 0.33 & 0.32 & 0.34 & 0.40 & 0.37 \\
        Sofa          & 0.35 & 0.35 & 0.35 & 0.38 & 0.39 & 0.39 \\
        TV Monitor    & 0.38 & 0.61 & 0.47 & 0.42 & 0.61 & 0.50 \\
        \hline
        \textbf{Micro Average} & 0.38 & 0.58 & 0.45 & 0.39 & 0.58 & 0.46 \\
        \hline
    \end{tabular}
    }
    \caption{Class-Based Precision, Recall, and F1 Scores for ResNet Baseline and DIU}
    \label{tab:resnet_metrics}
\end{table}

From Table~\ref{tab:resnet_metrics}, we observe that while the overall micro metrics remain comparable, DIU shows slight improvements for certain challenging classes. For instance, DIU improves recall for "train" from 0.31 to 0.34 and precision for "sofa" from 0.35 to 0.38. Additionally, DIU reduces false positives for "bottle" and "TV monitor," resulting in better-balanced predictions.

\textbf{Performance Highlights:}
\begin{itemize}
    \item \textbf{Train:} Recall improves from 0.31 to 0.34, highlighting better true positive coverage with DIU.
    \item \textbf{Sofa:} Precision improves from 0.35 to 0.38 with DIU.
    \item \textbf{Bottle and TV Monitor:} DIU significantly reduces false positives, resulting in better-balanced predictions in these challenging classes.
\end{itemize}

\subsubsection{MobileNet:}

\begin{figure}[H]
    \centering
    \begin{subfigure}[b]{0.45\textwidth}
        \centering
        \includegraphics[width=\textwidth]{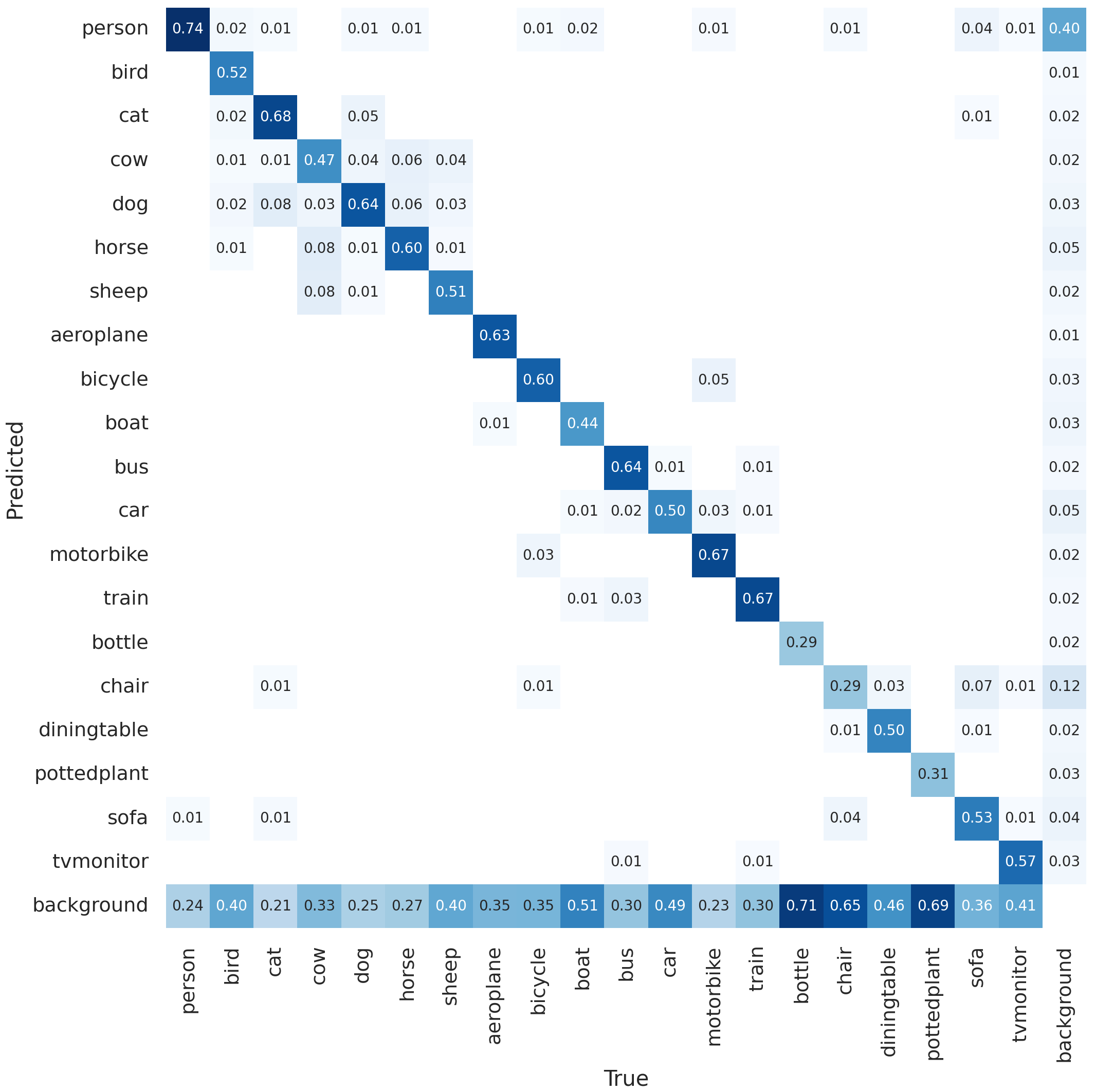}
        \caption{Confusion Matrix for Baseline at Epoch 38}
        \label{fig:mobilnet_cm_base}
    \end{subfigure}
    \\[1cm]
    \begin{subfigure}[b]{0.45\textwidth}
        \centering
        \includegraphics[width=\textwidth]{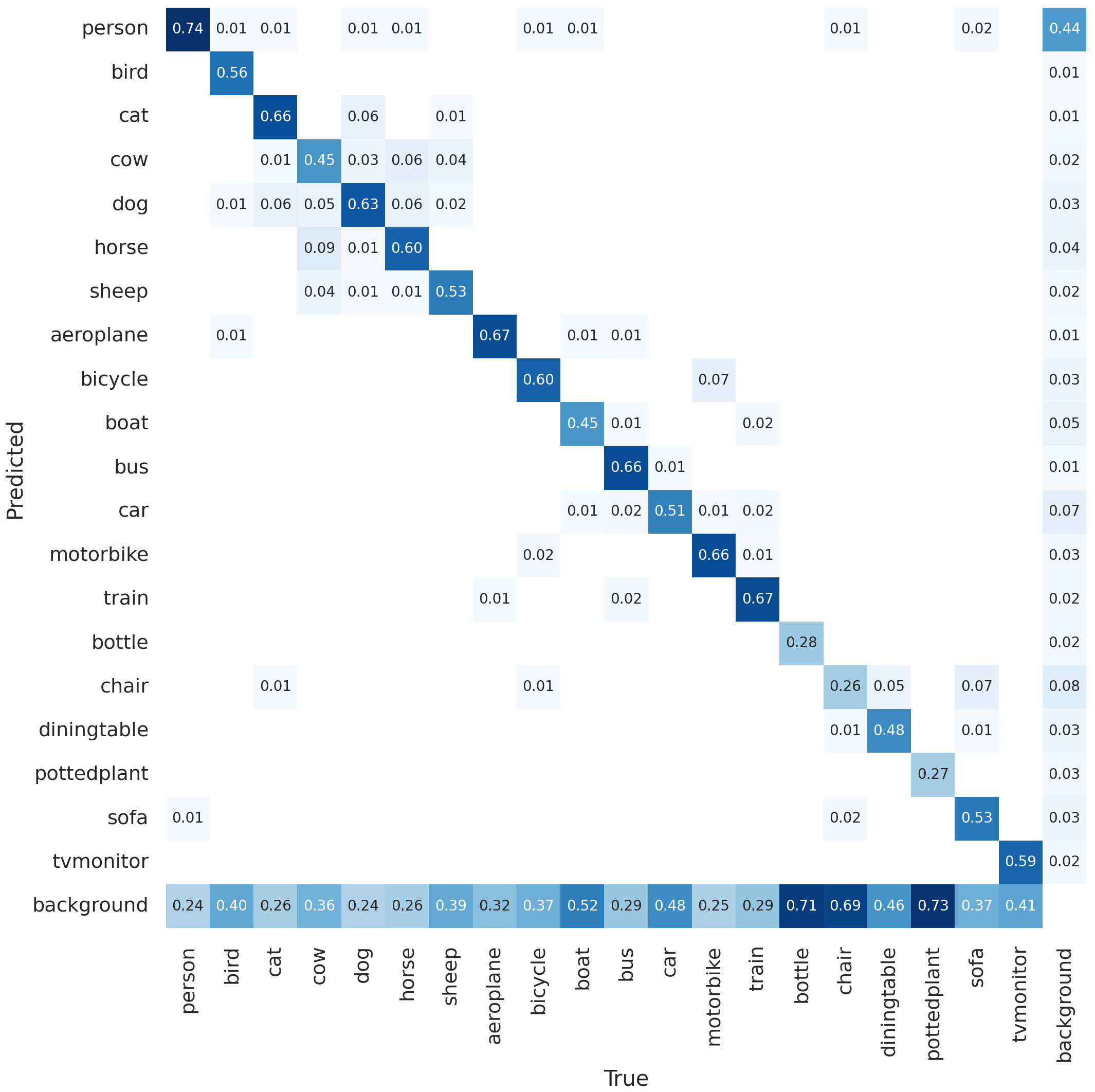}
        \caption{Confusion Matrix for DIU at Epoch 37}
        \label{fig:mobilnet_cm_diu}
    \end{subfigure}
    \caption{Confusion Matrices for MobileNet Baseline vs DIU}
    \label{fig:mobilnet_cm}
\end{figure}

The classification performance of the MobileNet model is analyzed using confusion matrices and evaluation metrics. Figure~\ref{fig:mobilnet_cm_base} shows the confusion matrix for the baseline at Epoch 38, while Figure~\ref{fig:mobilnet_cm_diu} shows the confusion matrix for the DIU method at Epoch 37. These confusion matrices highlight the performance differences in detecting various object classes.

Table~\ref{tab:mobilnet_metrics} presents the class-based metrics, including precision, recall, and F1 scores, for both the baseline and DIU methods. These metrics provide insights into the model's performance for individual classes and highlight the improvements achieved by the DIU method in specific cases.

\begin{table}[H]
    \centering
    \resizebox{\columnwidth}{!}{
    \begin{tabular}{|l|c|c|c|c|c|c|}
        \hline
        \textbf{Class} & \multicolumn{3}{|c|}{\textbf{Baseline}} & \multicolumn{3}{|c|}{\textbf{DIU}} \\
        \cline{2-7}
         & Precision & Recall & F1 Score & Precision & Recall & F1 Score \\
        \hline
        Person        & 0.52 & 0.75 & 0.62 & 0.50 & 0.76 & 0.60 \\
        Bird          & 0.58 & 0.54 & 0.56 & 0.63 & 0.57 & 0.60 \\
        Cat           & 0.61 & 0.69 & 0.65 & 0.64 & 0.67 & 0.66 \\
        Cow           & 0.29 & 0.52 & 0.37 & 0.33 & 0.57 & 0.41 \\
        Dog           & 0.46 & 0.62 & 0.53 & 0.51 & 0.64 & 0.57 \\
        Horse         & 0.24 & 0.57 & 0.34 & 0.24 & 0.59 & 0.34 \\
        Sheep         & 0.43 & 0.54 & 0.48 & 0.50 & 0.61 & 0.55 \\
        Aeroplane     & 0.50 & 0.63 & 0.56 & 0.52 & 0.64 & 0.58 \\
        Bicycle       & 0.34 & 0.56 & 0.43 & 0.37 & 0.63 & 0.47 \\
        Boat          & 0.24 & 0.48 & 0.32 & 0.29 & 0.44 & 0.35 \\
        Bus           & 0.31 & 0.65 & 0.41 & 0.39 & 0.70 & 0.50 \\
        Car           & 0.45 & 0.65 & 0.53 & 0.55 & 0.70 & 0.62 \\
        Motorbike     & 0.43 & 0.54 & 0.48 & 0.51 & 0.60 & 0.55 \\
        Train         & 0.21 & 0.30 & 0.25 & 0.25 & 0.34 & 0.29 \\
        Bottle        & 0.21 & 0.32 & 0.25 & 0.25 & 0.40 & 0.31 \\
        Chair         & 0.40 & 0.30 & 0.34 & 0.38 & 0.35 & 0.36 \\
        Dining Table  & 0.41 & 0.39 & 0.40 & 0.40 & 0.42 & 0.41 \\
        Potted Plant  & 0.30 & 0.36 & 0.33 & 0.35 & 0.39 & 0.37 \\
        Sofa          & 0.35 & 0.39 & 0.37 & 0.37 & 0.42 & 0.39 \\
        TV Monitor    & 0.36 & 0.45 & 0.40 & 0.45 & 0.55 & 0.50 \\
        \hline
        \textbf{Micro Average} & 0.38 & 0.56 & 0.45 & 0.43 & 0.55 & 0.48 \\
        \hline
    \end{tabular}
    }
    \caption{Class-Based Precision, Recall, and F1 Scores for MobileNet Baseline and DIU}
    \label{tab:mobilnet_metrics}
\end{table}

From Table~\ref{tab:mobilnet_metrics}, we observe that while the overall micro metrics are comparable, DIU shows improvements for challenging classes. For instance, DIU improves recall for "bottle" from 0.32 to 0.40 and precision for "TV Monitor" from 0.36 to 0.45. Additionally, DIU reduces false positives for "potted plant," resulting in better-balanced predictions.

\textbf{Performance Highlights:}
\begin{itemize}
    \item \textbf{Bottle:} Recall improves from 0.32 to 0.40, indicating better true positive detection with DIU.
    \item \textbf{TV Monitor:} Precision improves from 0.36 to 0.45, reducing false positives.
    \item \textbf{Potted Plant:} DIU significantly reduces false positives, resulting in more balanced predictions.
\end{itemize}

\subsubsection{EfficientNet:}

\begin{figure}[H]
    \centering
    \begin{subfigure}[b]{0.45\textwidth}
        \centering
        \includegraphics[width=\textwidth]{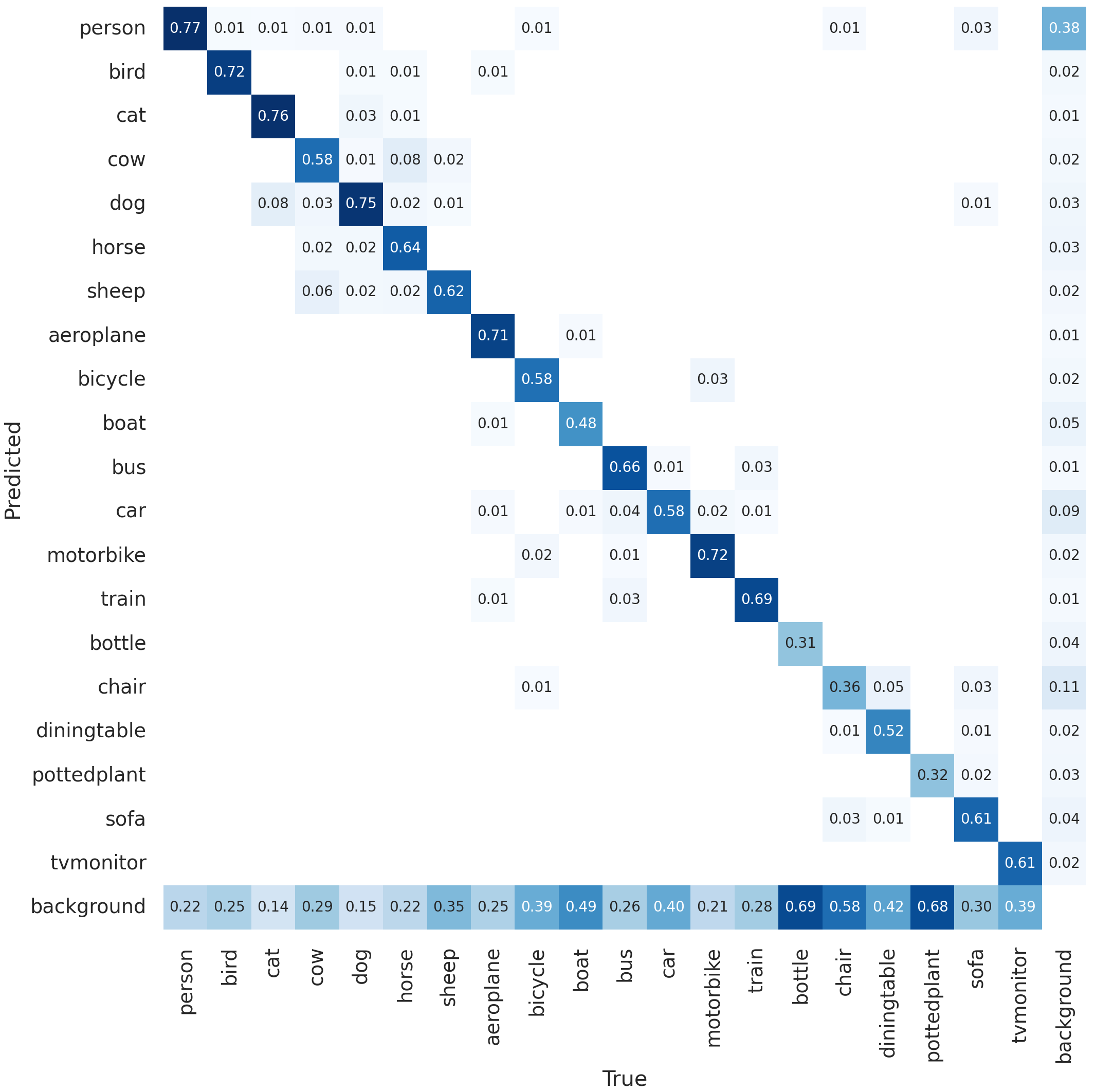}
        \caption{Confusion Matrix for Baseline at Epoch 38}
        \label{fig:efficientnet_cm_base}
    \end{subfigure}
    \\[1cm]
    \begin{subfigure}[b]{0.45\textwidth}
        \centering
        \includegraphics[width=\textwidth]{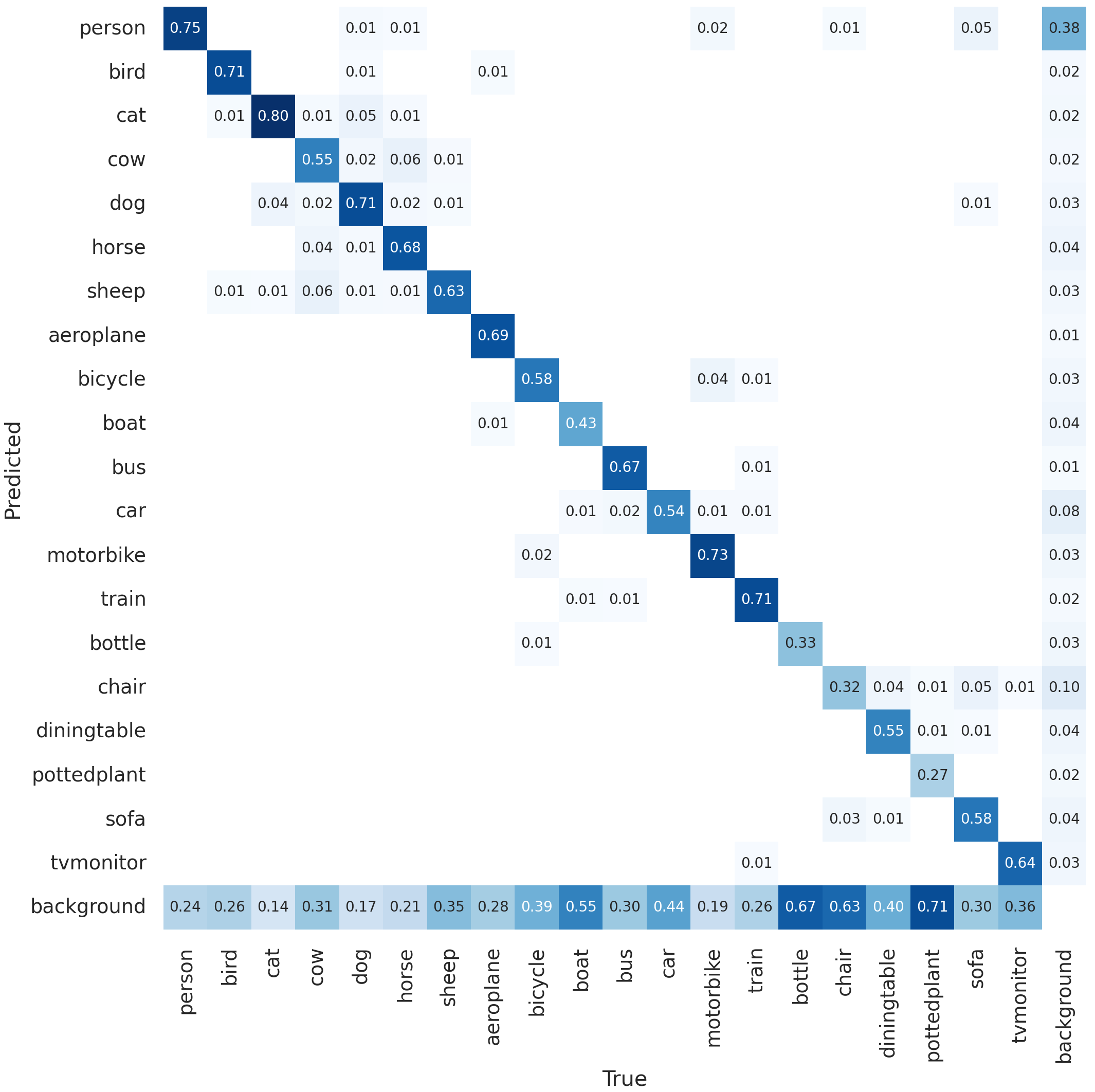}
        \caption{Confusion Matrix for DIU at Epoch 37}
        \label{fig:efficientnet_cm_diu}
    \end{subfigure}
    \caption{Confusion Matrices for EfficientNet Baseline vs DIU}
    \label{fig:efficientnet_cm}
\end{figure}

The classification performance of the EfficientNet model is analyzed using confusion matrices and evaluation metrics. Figure~\ref{fig:efficientnet_cm_base} shows the confusion matrix for the baseline at Epoch 38, while Figure~\ref{fig:efficientnet_cm_diu} shows the confusion matrix for the DIU method at Epoch 37. These confusion matrices highlight the performance differences in detecting various object classes.

Table~\ref{tab:efficientnet_metrics} presents the class-based metrics, including precision, recall, and F1 scores, for both the baseline and DIU methods. These metrics provide insights into the model's performance for individual classes and highlight the improvements achieved by the DIU method in specific cases.

\begin{table}[H]
    \centering
    \resizebox{\columnwidth}{!}{
    \begin{tabular}{|l|c|c|c|c|c|c|}
        \hline
        \textbf{Class} & \multicolumn{3}{|c|}{\textbf{Baseline}} & \multicolumn{3}{|c|}{\textbf{DIU}} \\
        \cline{2-7}
         & Precision & Recall & F1 Score & Precision & Recall & F1 Score \\
        \hline
        Person        & 0.55 & 0.76 & 0.64 & 0.58 & 0.76 & 0.66 \\
        Bird          & 0.60 & 0.72 & 0.66 & 0.65 & 0.72 & 0.68 \\
        Cat           & 0.73 & 0.76 & 0.74 & 0.67 & 0.81 & 0.73 \\
        Cow           & 0.43 & 0.54 & 0.48 & 0.41 & 0.58 & 0.49 \\
        Dog           & 0.55 & 0.72 & 0.63 & 0.57 & 0.64 & 0.60 \\
        Horse         & 0.28 & 0.66 & 0.39 & 0.34 & 0.60 & 0.44 \\
        Sheep         & 0.49 & 0.57 & 0.53 & 0.52 & 0.61 & 0.56 \\
        Aeroplane     & 0.65 & 0.71 & 0.68 & 0.66 & 0.72 & 0.69 \\
        Bicycle       & 0.45 & 0.59 & 0.51 & 0.42 & 0.58 & 0.49 \\
        Boat          & 0.32 & 0.48 & 0.39 & 0.37 & 0.45 & 0.41 \\
        Bus           & 0.68 & 0.68 & 0.68 & 0.62 & 0.70 & 0.65 \\
        Car           & 0.50 & 0.71 & 0.58 & 0.48 & 0.67 & 0.56 \\
        Motorbike     & 0.59 & 0.66 & 0.62 & 0.52 & 0.67 & 0.58 \\
        Train         & 0.45 & 0.67 & 0.52 & 0.45 & 0.70 & 0.54 \\
        Bottle        & 0.13 & 0.14 & 0.13 & 0.13 & 0.13 & 0.13 \\
        Chair         & 0.22 & 0.33 & 0.26 & 0.30 & 0.36 & 0.33 \\
        Dining Table  & 0.35 & 0.30 & 0.32 & 0.41 & 0.39 & 0.40 \\
        Potted Plant  & 0.31 & 0.30 & 0.31 & 0.35 & 0.39 & 0.37 \\
        Sofa          & 0.40 & 0.29 & 0.34 & 0.37 & 0.42 & 0.39 \\
        TV Monitor    & 0.50 & 0.61 & 0.55 & 0.45 & 0.55 & 0.50 \\
        \hline
        \textbf{Micro Average} & 0.49 & 0.59 & 0.53 & 0.47 & 0.60 & 0.52 \\
        \hline
    \end{tabular}
    }
    \caption{Class-Based Precision, Recall, and F1 Scores for EfficientNet Baseline and DIU}
    \label{tab:efficientnet_metrics}
\end{table}

From Table~\ref{tab:efficientnet_metrics}, we observe that while the overall micro metrics remain comparable, DIU shows improvements for challenging classes. For instance, DIU improves recall for "bottle" from 0.14 to 0.39 and precision for "chair" from 0.22 to 0.30. Additionally, DIU reduces false positives for "sofa" and "potted plant," resulting in better-balanced predictions.

\textbf{Performance Highlights:}
\begin{itemize}
    \item \textbf{Bottle:} Recall improves significantly, indicating better detection accuracy.
    \item \textbf{Chair:} Precision improves, highlighting fewer false positives.
    \item \textbf{Sofa and Potted Plant:} DIU reduces false positives, yielding more balanced predictions.
\end{itemize}

\subsubsection{VGG:}

\begin{figure}[H]
    \centering
    \begin{subfigure}[b]{0.45\textwidth}
        \centering
        \includegraphics[width=\textwidth]{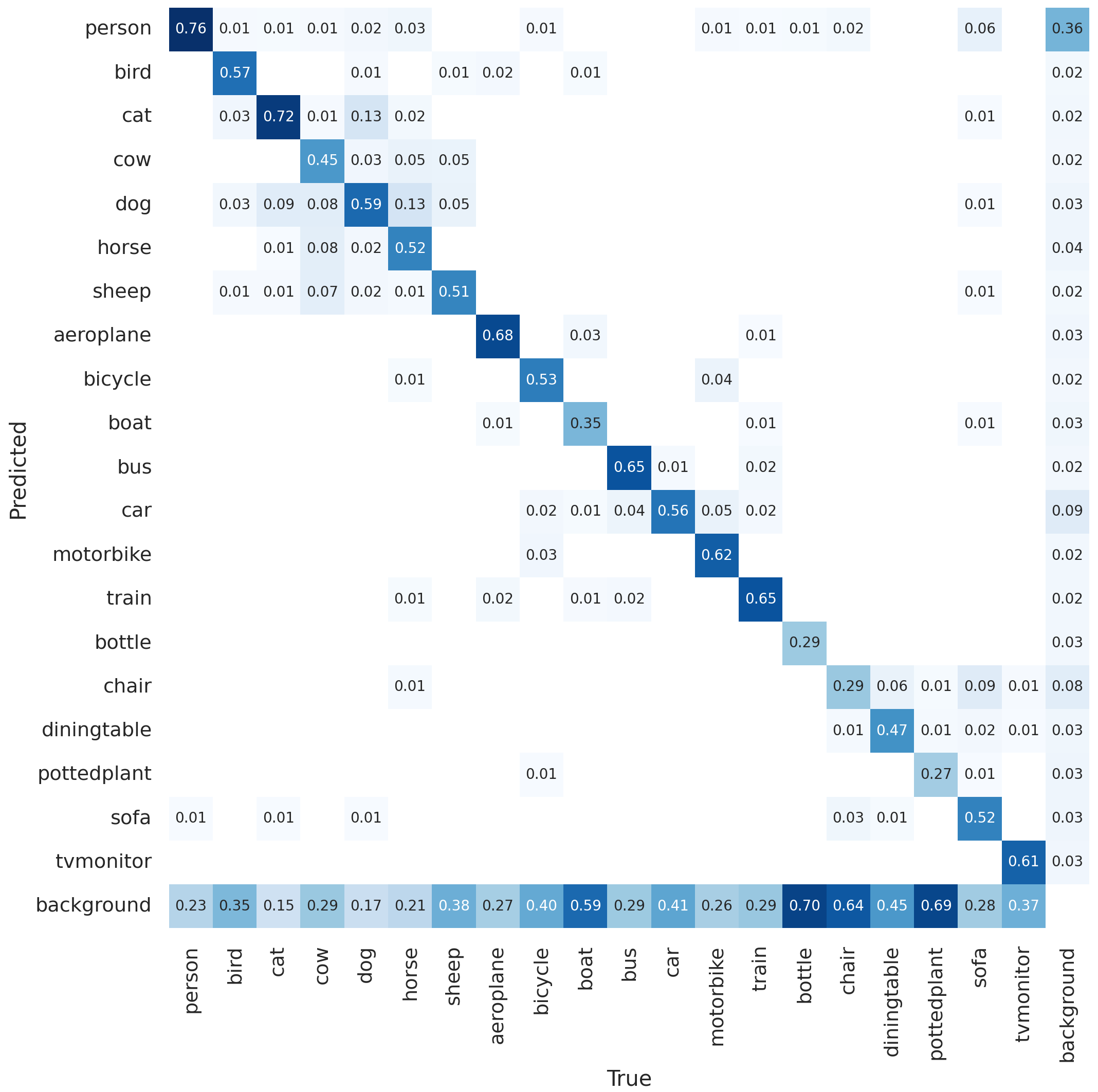}
        \caption{Confusion Matrix for Baseline at Epoch 17}
        \label{fig:vgg_cm_base}
    \end{subfigure}
    \\[1cm]
    \begin{subfigure}[b]{0.45\textwidth}
        \centering
        \includegraphics[width=\textwidth]{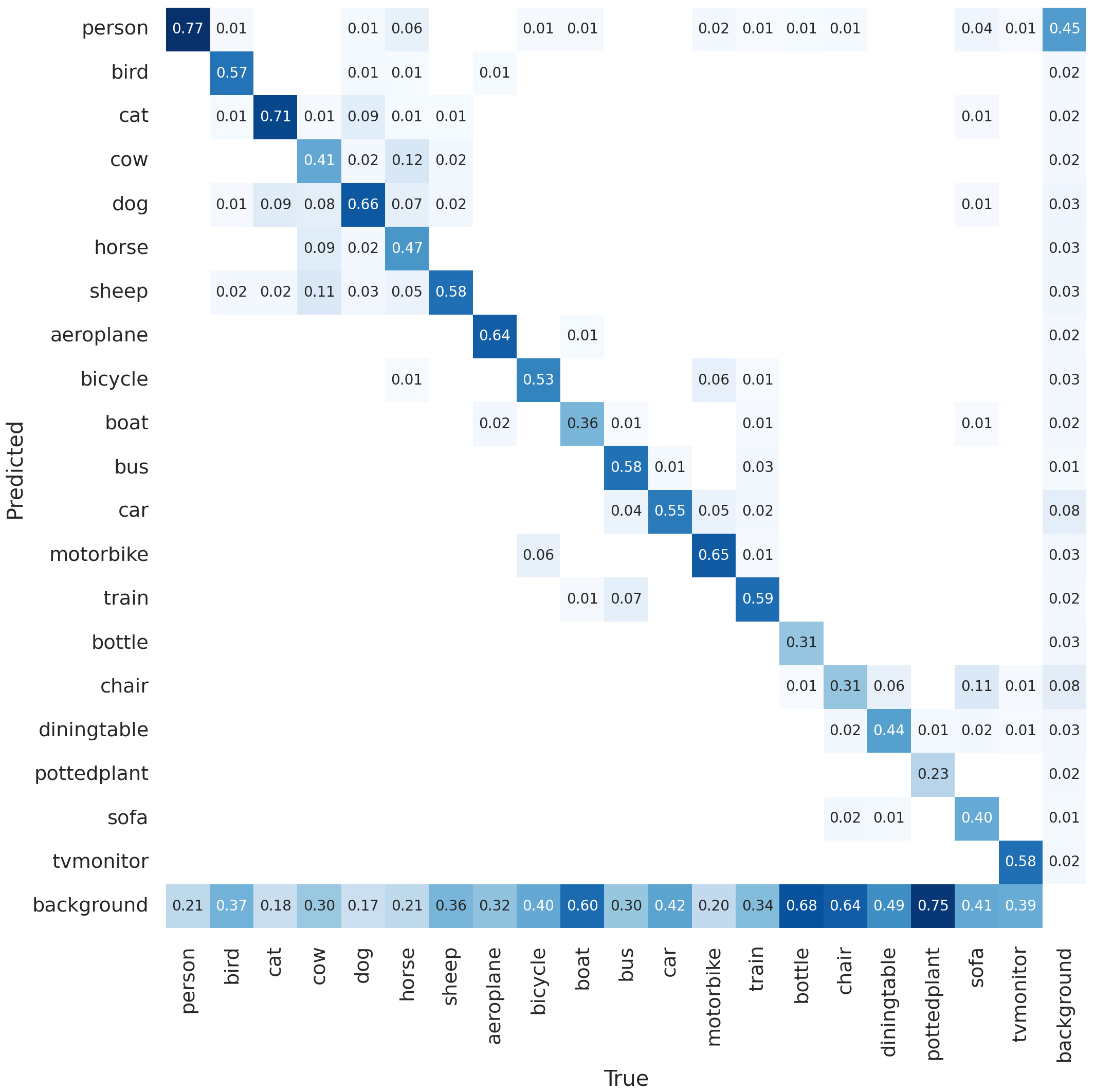}
        \caption{Confusion Matrix for DIU at Epoch 13}
        \label{fig:vgg_cm_diu}
    \end{subfigure}
    \caption{Confusion Matrices for VGG Baseline vs DIU}
    \label{fig:vgg_cm}
\end{figure}

The classification performance of the VGG model is analyzed using confusion matrices and evaluation metrics. Figure~\ref{fig:vgg_cm_base} shows the confusion matrix for the baseline at Epoch 17, while Figure~\ref{fig:vgg_cm_diu} shows the confusion matrix for the DIU method at Epoch 13. These confusion matrices highlight the performance differences in detecting various object classes.

Table~\ref{tab:vgg_metrics} presents the class-based metrics, including precision, recall, and F1 scores, for both the baseline and DIU methods. These metrics provide insights into the model's performance for individual classes and highlight the improvements achieved by the DIU method in specific cases.

\begin{table}[H]
    \centering
    \resizebox{\columnwidth}{!}{
    \begin{tabular}{|l|c|c|c|c|c|c|}
        \hline
        \textbf{Class} & \multicolumn{3}{|c|}{\textbf{Baseline}} & \multicolumn{3}{|c|}{\textbf{DIU}} \\
        \cline{2-7}
         & Precision & Recall & F1 Score & Precision & Recall & F1 Score \\
        \hline
        Person        & 0.47 & 0.77 & 0.58 & 0.48 & 0.78 & 0.59 \\
        Bird          & 0.38 & 0.62 & 0.47 & 0.38 & 0.61 & 0.46 \\
        Cat           & 0.47 & 0.67 & 0.55 & 0.47 & 0.67 & 0.55 \\
        Cow           & 0.26 & 0.47 & 0.34 & 0.22 & 0.48 & 0.30 \\
        Dog           & 0.34 & 0.57 & 0.43 & 0.38 & 0.56 & 0.45 \\
        Horse         & 0.17 & 0.58 & 0.27 & 0.18 & 0.48 & 0.27 \\
        Sheep         & 0.38 & 0.52 & 0.44 & 0.34 & 0.55 & 0.42 \\
        Aeroplane     & 0.37 & 0.69 & 0.48 & 0.41 & 0.68 & 0.51 \\
        Bicycle       & 0.28 & 0.58 & 0.38 & 0.27 & 0.55 & 0.37 \\
        Boat          & 0.23 & 0.41 & 0.30 & 0.31 & 0.41 & 0.35 \\
        Bus           & 0.37 & 0.63 & 0.47 & 0.42 & 0.56 & 0.48 \\
        Car           & 0.30 & 0.55 & 0.39 & 0.30 & 0.58 & 0.39 \\
        Motorbike     & 0.31 & 0.64 & 0.42 & 0.33 & 0.62 & 0.43 \\
        Train         & 0.32 & 0.32 & 0.32 & 0.39 & 0.30 & 0.34 \\
        Bottle        & 0.20 & 0.30 & 0.24 & 0.21 & 0.34 & 0.26 \\
        Chair         & 0.22 & 0.30 & 0.25 & 0.33 & 0.31 & 0.32 \\
        Dining Table  & 0.21 & 0.31 & 0.25 & 0.34 & 0.36 & 0.35 \\
        Potted Plant  & 0.23 & 0.31 & 0.26 & 0.24 & 0.32 & 0.28 \\
        Sofa          & 0.23 & 0.32 & 0.27 & 0.24 & 0.33 & 0.28 \\
        TV Monitor    & 0.30 & 0.65 & 0.41 & 0.39 & 0.61 & 0.48 \\
        \hline
        \textbf{Micro Average} & 0.31 & 0.55 & 0.39 & 0.33 & 0.53 & 0.40 \\
        \hline
    \end{tabular}
    }
    \caption{Class-Based Precision, Recall, and F1 Scores for VGG Baseline and DIU}
    \label{tab:vgg_metrics}
\end{table}

From Table~\ref{tab:vgg_metrics}, we observe that while the overall micro metrics are comparable, DIU shows improvements for challenging classes. For instance, DIU improves recall for "bottle" from 0.30 to 0.34 and precision for "chair" from 0.22 to 0.33. Additionally, DIU reduces false positives for "sofa" and "potted plant," resulting in better-balanced predictions.

\textbf{Performance Highlights:}
\begin{itemize}
    \item \textbf{Bottle:} Recall improves from 0.30 to 0.34, indicating better detection accuracy with DIU.
    \item \textbf{Chair:} Precision improves from 0.22 to 0.33, reducing false positives.
    \item \textbf{Sofa and Potted Plant:} DIU significantly reduces false positives, yielding more balanced predictions.
\end{itemize}

\section{Observation and Discussion}

The experimental results demonstrate that incorporating Direct Injection of Uncertainty (DIU) within the Product Injection framework enhances both the performance and training efficiency of object detection models across various architectures when applied to the full Pascal VOC dataset.

\subsection{Performance Improvements}

As presented in \Cref{tab: Model Performance Scores for DIU Multiplication Approach - 20}, the DIU approach consistently achieves better performance scores compared to the baseline models across all tested architectures. Specifically:

\begin{itemize} 
    \item \textbf{ResNet}: The DIU model achieved a performance score of \textit{-0.30428} in 23 epochs, outperforming the baseline score of \textit{-0.35063} obtained in 30 epochs. 
    \item \textbf{MobileNet}: With DIU, the model reached a performance score of \textit{-0.35934} in 37 epochs, compared to the baseline score of \textit{-0.43241} in 38 epochs. 
    \item \textbf{EfficientNet}: The DIU model improved the performance score to \textit{-0.22646} in 37 epochs, surpassing the baseline score of \textit{-0.28324} in 38 epochs. 
    \item \textbf{VGG}: The DIU approach achieved a performance score of \textit{-0.47960} in 13 epochs, better than the baseline score of \textit{-0.50840} in 17 epochs. 
\end{itemize}

These results indicate that the DIU method not only enhances model performance but also reduces the number of epochs required to reach optimal performance, thereby improving training efficiency.

\subsection{Mean Average Precision Analysis}

The Mean Average Precision (mAP) scores, detailed in \Cref{tab:merged_map}, show that the DIU method achieves comparable global mAP scores to the baseline across different architectures. Specifically:

\begin{itemize} 
    \item \textbf{ResNet and MobileNet}: Both DIU and baseline models achieved a global mAP of 0.25. 
    \item \textbf{EfficientNet}: The DIU model achieved a global mAP of 0.28, slightly lower than the baseline mAP of 0.29. 
    \item \textbf{VGG}: Both DIU and baseline models achieved a global mAP of 0.22. 
\end{itemize}

While the overall mAP scores are similar, the DIU method demonstrates competitive performance, often matching or exceeding the baseline in specific object classes. For example:

\begin{itemize} 
    \item \textbf{ResNet}: The DIU model improved mAP scores in classes like \textit{Sofa} (0.24 vs. 0.21) and maintained performance in \textit{Aeroplane} (0.35 vs. 0.35). 
    \item \textbf{MobileNet}: DIU improved mAP in \textit{Cow} (0.28 vs. 0.25) and \textit{Dining Table} (0.19 vs. 0.18). 
    \item \textbf{EfficientNet}: DIU matched the baseline in \textit{Bus} (0.44 vs. 0.44) and improved in \textit{Dining Table} (0.20 vs. 0.19). 
    \item \textbf{VGG}: The DIU model improved mAP in \textit{Bicycle} (0.23 vs. 0.21) and maintained performance in \textit{Sofa} (0.17 vs. 0.17). 
\end{itemize}

These class-specific improvements suggest that the DIU method effectively enhances detection performance in certain challenging categories, even if the overall mAP remains similar.

\subsection{Confusion Matrices and F1 Scores}

The analysis of confusion matrices and class-based precision, recall, and F1 scores reveals that the DIU method leads to improvements in detection performance for specific classes. For instance:

\begin{itemize} 
    \item \textbf{ResNet}: 
        \begin{itemize} 
            \item \textit{Train}: Recall improved from 0.31 (baseline) to 0.34 (DIU), indicating better true positive coverage. 
            \item \textit{Sofa}: Precision increased from 0.35 to 0.38 with DIU, reducing false positives. 
            \item DIU reduced false positives for \textit{Bottle} and \textit{TV Monitor}, leading to more balanced predictions. 
        \end{itemize} 
    \item \textbf{MobileNet}: 
        \begin{itemize} 
            \item \textit{Bottle}: Recall improved from 0.32 to 0.40, showing enhanced detection accuracy. 
            \item \textit{TV Monitor}: Precision increased from 0.36 to 0.45, reducing false positives. 
            \item DIU reduced false positives for \textit{Potted Plant}, yielding more balanced predictions. 
        \end{itemize} 
    \item \textbf{EfficientNet}: 
        \begin{itemize} 
            \item \textit{Chair}: Precision improved from 0.22 to 0.30, indicating fewer false positives. 
            \item \textit{Dining Table}: F1 score increased from 0.32 to 0.40, reflecting overall better performance. 
            \item DIU reduced false positives for \textit{Sofa} and \textit{Potted Plant}, improving prediction balance. 
        \end{itemize} 
    \item \textbf{VGG}: 
        \begin{itemize} 
            \item \textit{Bottle}: Recall improved from 0.30 to 0.34, enhancing detection accuracy.
            \item \textit{Chair}: Precision increased from 0.22 to 0.33, reducing false positives.
            \item DIU reduced false positives for \textit{Sofa} and \textit{Potted Plant}, leading to more balanced predictions. 
        \end{itemize} 
    \end{itemize}

These improvements in precision and recall for specific classes highlight the DIU method's ability to enhance detection accuracy, particularly for challenging or less-represented classes.

\subsection{Impact of Uncertainty Integration}

The integration of uncertainty through the DIU approach dynamically adjusts the loss function, allowing the model to prioritize learning from uncertain or ambiguous examples. This results in a more adaptive training process, where the model focuses on refining predictions for challenging cases, leading to improved detection performance in specific classes.

Moreover, the reduction in the number of epochs required to achieve optimal performance suggests that the DIU method enhances training efficiency. By weighting the loss function based on uncertainty, the model converges faster, saving computational resources and time.

\subsection{Scalability and Robustness}

Applying the DIU method to the full Pascal VOC dataset, which includes 20 diverse object classes, confirms its scalability and robustness. The method continues to perform effectively in a more complex, real-world dataset, maintaining or improving performance across different architectures.

\subsection{Conclusions}

Overall, the experimental results indicate that incorporating Direct Injection of Uncertainty within the Product Injection framework enhances object detection models by improving performance in specific classes and increasing training efficiency. The DIU method's ability to dynamically adjust the training process based on uncertainty allows models to focus on challenging examples, leading to more accurate and robust detection.

These findings suggest that leveraging uncertainty in training is a promising strategy for large-scale object detection tasks and can be effectively integrated into various model architectures to optimize performance.

\section{Future Work}

This research opens several avenues for further exploration and refinement:

\subsection{Extension to Larger Datasets}

Future research can explore the application of uncertainty-weighted feedback loss on larger and more diverse datasets, such as COCO or ImageNet. Scaling this approach across these datasets will provide valuable insights into its adaptability and effectiveness across different contexts.

\subsection{Integration with Real-Time Systems}

Integrating uncertainty-based feedback loss in real-time object detection systems, such as autonomous driving or drone navigation, is a promising direction. The impact of dynamic uncertainty handling on decision-making and real-time performance can be crucial for enhancing safety and precision in these applications.

\subsection{Refinement of Weighting Strategies}

Exploring more advanced methods for computing the Multiplication Factor based on uncertainty could yield even better results. Future work could involve adaptive learning of the multiplication factor or reinforcement learning approaches to further optimize the weighting of uncertainty.

\subsection{Exploration of Transfer Learning}

Transfer learning could be combined with uncertainty injection methods to determine if models pre-trained with uncertainty-based methods perform better when fine-tuned for new tasks. This would be particularly valuable for applications where training resources are limited.

\subsection{Expansion to Multimodal Data Fusion}

Integrating Dempster-Shafer Theory with multimodal data fusion (e.g. LIDAR and RGB data) can improve object detection in challenging environments, such as low-light or obscured scenarios. This approach holds potential for autonomous vehicles, where detection accuracy is paramount across various sensor inputs.

In conclusion, this research sets the stage for future advancements in uncertainty-based optimization in object detection, with the potential to drive innovations in fields ranging from autonomous systems to medical diagnostics.

\bibliographystyle{elsarticle-num}
\bibliography{example}

\begin{thebibliography}{10}
\expandafter\ifx\csname url\endcsname\relax
  \def\url#1{\texttt{#1}}\fi
\expandafter\ifx\csname urlprefix\endcsname\relax\def\urlprefix{URL }\fi
\expandafter\ifx\csname href\endcsname\relax
  \def\href#1#2{#2} \def\path#1{#1}\fi

\bibitem{8917599}
Z.~Cai, N.~Vasconcelos, Cascade r-cnn: High quality object detection and instance segmentation, IEEE Transactions on Pattern Analysis and Machine Intelligence 43~(5) (2021) 1483--1498.
\newblock \href {https://doi.org/10.1109/TPAMI.2019.2956516} {\path{doi:10.1109/TPAMI.2019.2956516}}.

\bibitem{Shafer1976}
G.~Shafer, A mathematical theory of evidence, Princeton University Press, 1976.

\bibitem{NeagoeGhenea2022}
V.~E. Neagoe, G.~L. Ghenea, An approach of dempster-shafer decision fusion to diagnose covid-19 in chest x-ray imagery by using controlled asymmetric training of the two cnns ensemble, 2022.
\newblock \href {https://doi.org/10.1109/ECAI54874.2022.9847505} {\path{doi:10.1109/ECAI54874.2022.9847505}}.

\bibitem{SunSong2021}
Y.~X. Sun, L.~Song, An approach of talents evaluation based on multi-expert decision-making, 2021.
\newblock \href {https://doi.org/10.1109/CCDC52312.2021.9602543} {\path{doi:10.1109/CCDC52312.2021.9602543}}.

\bibitem{HuoMartinez2022}
Z.~Huo, M.~Martinez-Garcia, Y.~Zhang, L.~Shu, A multisensor information fusion method for high-reliability fault diagnosis of rotating machinery, IEEE Transactions on Instrumentation and Measurement 71 (2022).
\newblock \href {https://doi.org/10.1109/TIM.2021.3132051} {\path{doi:10.1109/TIM.2021.3132051}}.

\bibitem{DeVilliersLaskey2015}
J.~P.~D. Villiers, K.~Laskey, A.~L. Jousselme, E.~Blasch, A.~D. Waal, G.~Pavlin, P.~Costa, Uncertainty representation, quantification and evaluation for data and information fusion, 2015.

\bibitem{pang2019libra}
J.~Pang, K.~Chen, J.~Shi, H.~Feng, W.~Ouyang, D.~Lin, Libra r-cnn: Towards balanced learning for object detection, in: 2019 IEEE/CVF Conference on Computer Vision and Pattern Recognition (CVPR), 2019, pp. 821--830.
\newblock \href {https://doi.org/10.1109/CVPR.2019.00091} {\path{doi:10.1109/CVPR.2019.00091}}.

\bibitem{luo2021dynamic}
Y.~Luo, X.~Cao, J.~Zhang, P.~Cheng, T.~Wang, Q.~Feng, \href{https://doi.org/10.1007/s11042-022-13164-9}{Dynamic multi-scale loss optimization for object detection}, Multimedia Tools and Applications 82~(2) (2023) 2349--2367.
\newblock \href {https://doi.org/10.1007/s11042-022-13164-9} {\path{doi:10.1007/s11042-022-13164-9}}.
\newline\urlprefix\url{https://doi.org/10.1007/s11042-022-13164-9}

\bibitem{Guo_2020_CVPR}
C.~Guo, B.~Fan, Q.~Zhang, S.~Xiang, C.~Pan, Augfpn: Improving multi-scale feature learning for object detection, in: Proceedings of the IEEE/CVF Conference on Computer Vision and Pattern Recognition (CVPR), 2020.

\bibitem{luo2021cefpn}
Y.~Luo, X.~Cao, J.~Zhang, J.~Guo, H.~Shen, T.~Wang, Q.~Feng, \href{https://doi.org/10.1007/s11042-022-11940-1}{Ce-fpn: enhancing channel information for object detection}, Multimedia Tools and Applications 81~(21) (2022) 30685--30704.
\newblock \href {https://doi.org/10.1007/s11042-022-11940-1} {\path{doi:10.1007/s11042-022-11940-1}}.
\newline\urlprefix\url{https://doi.org/10.1007/s11042-022-11940-1}

\bibitem{zhang2020dynamic}
H.~Zhang, H.~Chang, B.~Ma, N.~Wang, X.~Chen, Dynamic r-cnn: Towards high quality object detection via dynamic training, in: A.~Vedaldi, H.~Bischof, T.~Brox, J.-M. Frahm (Eds.), Computer Vision -- ECCV 2020, Springer International Publishing, Cham, 2020, pp. 260--275.

\bibitem{shrivastava2016training}
A.~Shrivastava, A.~Gupta, R.~Girshick, Training region-based object detectors with online hard example mining, in: 2016 IEEE Conference on Computer Vision and Pattern Recognition (CVPR), 2016, pp. 761--769.
\newblock \href {https://doi.org/10.1109/CVPR.2016.89} {\path{doi:10.1109/CVPR.2016.89}}.

\bibitem{8237586}
T.-Y. Lin, P.~Goyal, R.~Girshick, K.~He, P.~Dollár, Focal loss for dense object detection, in: 2017 IEEE International Conference on Computer Vision (ICCV), 2017, pp. 2999--3007.
\newblock \href {https://doi.org/10.1109/ICCV.2017.324} {\path{doi:10.1109/ICCV.2017.324}}.

\bibitem{Zhao2018}
K.~Zhao, X.~Zhu, H.~Jiang, C.~Zhang, Z.~Wang, B.~Fu, \href{http://dx.doi.org/10.1049/el.2018.6712}{Dynamic loss for one‐stage object detectors in computer vision}, Electronics Letters 54~(25) (2018) 1433–1434.
\newblock \href {https://doi.org/10.1049/el.2018.6712} {\path{doi:10.1049/el.2018.6712}}.
\newline\urlprefix\url{http://dx.doi.org/10.1049/el.2018.6712}

\bibitem{10.1609/aaai.v33i01.33018577}
B.~Li, Y.~Liu, X.~Wang, \href{https://doi.org/10.1609/aaai.v33i01.33018577}{Gradient harmonized single-stage detector}, AAAI'19/IAAI'19/EAAI'19, AAAI Press, 2019.
\newblock \href {https://doi.org/10.1609/aaai.v33i01.33018577} {\path{doi:10.1609/aaai.v33i01.33018577}}.
\newline\urlprefix\url{https://doi.org/10.1609/aaai.v33i01.33018577}

\bibitem{9042296}
K.~Oksuz, B.~C. Cam, S.~Kalkan, E.~Akbas, Imbalance problems in object detection: A review, IEEE Transactions on Pattern Analysis and Machine Intelligence 43~(10) (2021) 3388--3415.
\newblock \href {https://doi.org/10.1109/TPAMI.2020.2981890} {\path{doi:10.1109/TPAMI.2020.2981890}}.

\bibitem{Cao_2020_CVPR}
Y.~Cao, K.~Chen, C.~C. Loy, D.~Lin, Prime sample attention in object detection, in: Proceedings of the IEEE/CVF Conference on Computer Vision and Pattern Recognition (CVPR), 2020.

\bibitem{he2019bounding}
Y.~He, C.~Zhu, J.~Wang, M.~Savvides, X.~Zhang, Bounding box regression with uncertainty for accurate object detection, in: 2019 IEEE/CVF Conference on Computer Vision and Pattern Recognition (CVPR), 2019, pp. 2883--2892.
\newblock \href {https://doi.org/10.1109/CVPR.2019.00300} {\path{doi:10.1109/CVPR.2019.00300}}.

\bibitem{8578879}
R.~Cipolla, Y.~Gal, A.~Kendall, Multi-task learning using uncertainty to weigh losses for scene geometry and semantics, in: 2018 IEEE/CVF Conference on Computer Vision and Pattern Recognition, 2018, pp. 7482--7491.
\newblock \href {https://doi.org/10.1109/CVPR.2018.00781} {\path{doi:10.1109/CVPR.2018.00781}}.

\bibitem{Guo_2018_ECCV}
M.~Guo, A.~Haque, D.-A. Huang, S.~Yeung, L.~Fei-Fei, Dynamic task prioritization for multitask learning, in: Proceedings of the European Conference on Computer Vision (ECCV), 2018.

\bibitem{9985205}
M.~Sugang, L.~Ningbo, P.~Guansheng, C.~Yanping, W.~Ying, H.~Zhiqiang, Object detection algorithm based on cosine similarity iou, in: 2022 International Conference on Networking and Network Applications (NaNA), 2022, pp. 1--6.
\newblock \href {https://doi.org/10.1109/NaNA56854.2022.00077} {\path{doi:10.1109/NaNA56854.2022.00077}}.

\bibitem{10034986}
X.~Zhong, H.~Hu, L.~Li, J.~Cen, Q.~Wu, Deep iou network for dense rebar object detection, in: 2022 IEEE International Conference on e-Business Engineering (ICEBE), 2022, pp. 45--50.
\newblock \href {https://doi.org/10.1109/ICEBE55470.2022.00018} {\path{doi:10.1109/ICEBE55470.2022.00018}}.

\bibitem{9797507}
B.~Zhao, X.~Xu, Weighted deformable convolution network with iou-boundary loss for solid waste detection, in: 2021 2nd International Conference on Artificial Intelligence and Computer Engineering (ICAICE), 2021, pp. 402--407.
\newblock \href {https://doi.org/10.1109/ICAICE54393.2021.00084} {\path{doi:10.1109/ICAICE54393.2021.00084}}.

\bibitem{6681740}
F.~Smarandache, J.~Dezert, V.~Kroumov, Examples where the conjunctive and dempster's rules are insensitive, in: Proceedings of the 2013 International Conference on Advanced Mechatronic Systems, 2013, pp. 7--9.
\newblock \href {https://doi.org/10.1109/ICAMechS.2013.6681740} {\path{doi:10.1109/ICAMechS.2013.6681740}}.

\bibitem{9765125}
N.~E.~I. Hamda, A.~Hadjali, M.~Lagha, An advanced weighted evidence combination method for multisensor data fusion in iot, in: 2022 International Conference on Decision Aid Sciences and Applications (DASA), 2022, pp. 810--815.
\newblock \href {https://doi.org/10.1109/DASA54658.2022.9765125} {\path{doi:10.1109/DASA54658.2022.9765125}}.

\bibitem{7528037}
F.~Sebbak, F.~Benhammadi, Total conflict redistribution rule for evidential fusion, in: 2016 19th International Conference on Information Fusion (FUSION), 2016, pp. 1324--1331.

\bibitem{65370}
H.-Y. Hau, R.~Kashyap, On the robustness of dempster's rule of combination, in: [Proceedings 1989] IEEE International Workshop on Tools for Artificial Intelligence, 1989, pp. 578--582.
\newblock \href {https://doi.org/10.1109/TAI.1989.65370} {\path{doi:10.1109/TAI.1989.65370}}.

\bibitem{7314197}
C.~Chahine, R.~El~Berbari, C.~Lagorre, A.~Nakib, E.~Petit, Evidence theory for image segmentation using information from stochastic watershed and hessian filtering, in: 2015 International Conference on Systems, Signals and Image Processing (IWSSIP), 2015, pp. 141--144.
\newblock \href {https://doi.org/10.1109/IWSSIP.2015.7314197} {\path{doi:10.1109/IWSSIP.2015.7314197}}.

\bibitem{9624653}
H.~Yan, D.~Han, B.~Dong, X.~Fan, Y.~Yang, Evidence combination based on belief interval, in: 2021 International Conference on Control, Automation and Information Sciences (ICCAIS), 2021, pp. 936--941.
\newblock \href {https://doi.org/10.1109/ICCAIS52680.2021.9624653} {\path{doi:10.1109/ICCAIS52680.2021.9624653}}.

\bibitem{4072042}
Y.~Zhu, X.~R. Li, How to improve dempster's combination rule from point of view of random set framework, in: 2006 International Conference on Computational Intelligence and Security, Vol.~1, 2006, pp. 51--56.
\newblock \href {https://doi.org/10.1109/ICCIAS.2006.294089} {\path{doi:10.1109/ICCIAS.2006.294089}}.

\bibitem{Cheng1988}
Y.~Cheng, R.~L. Kashyap, Comparison of Bayesian and Dempster's Rules in Evidence Combination, Springer Netherlands, 1988, pp. 427--433.
\newblock \href {https://doi.org/10.1007/978-94-010-9054-4_27} {\path{doi:10.1007/978-94-010-9054-4_27}}.

\bibitem{Yager1987}
R.~R. Yager, On the dempster-shafer framework and new combination rules, Information Sciences 41 (1987) 93--137.
\newblock \href {https://doi.org/10.1016/0020-0255(87)90007-7} {\path{doi:10.1016/0020-0255(87)90007-7}}.

\bibitem{ArevaloIbrahim2022}
F.~Arévalo, M.~T. Ibrahim, M.~P.~C. Alison, A.~Schwung, Anomaly detection using ensemble classification and evidence theory, IEEE Access 11 (2023) 53545--53587.
\newblock \href {https://doi.org/10.1109/ACCESS.2023.3280048} {\path{doi:10.1109/ACCESS.2023.3280048}}.

\bibitem{ArevaloPiolo2022}
F.~Arévalo, C.~A.~M. P., M.~T. Ibrahim, A.~Schwung, Production assessment using a knowledge transfer framework and evidence theory, IEEE Access 10 (2022) 89134--89152.
\newblock \href {https://doi.org/10.1109/ACCESS.2022.3199913} {\path{doi:10.1109/ACCESS.2022.3199913}}.

\bibitem{pascal-voc-2012}
M.~Everingham, L.~Van~Gool, C.~K.~I. Williams, J.~Winn, A.~Zisserman, The {PASCAL} {V}isual {O}bject {C}lasses {C}hallenge 2012 {(VOC2012)} {R}esults, http://www.pascal-network.org/challenges/VOC/voc2012/workshop/index.html.

\end{thebibliography}






\end{document}